\documentclass[3p,times,procedia]{elsarticle}

\usepackage{ecrc}

\volume{00}

\firstpage{1}

\journalname{Procedia Computer Science}

\runauth{G. Wang et al.}

\jid{procs}

\jnltitlelogo{Procedia Computer Science}

\CopyrightLine{2023}{Published by Elsevier Ltd.}

\usepackage{graphicx}
\DeclareGraphicsExtensions{.pdf,.jpeg,.png}
\usepackage{epstopdf}
\usepackage{txfonts}

\usepackage{amsmath}
\usepackage{amsthm}
\usepackage{amsfonts}
\usepackage{amssymb}
\usepackage{cancel}
\usepackage{enumitem}
\usepackage{thmtools}
\usepackage{thm-restate}

\newtheorem{remark}{Remark}%
\newtheorem{definition}{Definition}

\declaretheorem[name=Lemma]{lemma}

\usepackage{algorithmic}
\usepackage[ruled,norelsize]{algorithm2e}
\usepackage{multirow} 
\usepackage{xcolor}
\usepackage{setspace}
\makeatletter
\newcommand{\removelatexerror}{\let\@latex@error\@gobble}
\makeatother

\usepackage{array}

\usepackage{booktabs}

\usepackage[caption=false,font=normalsize,labelfont=sf,textfont=sf]{subfig}
\usepackage{url}

\usepackage{footnote}
\usepackage{threeparttable}

\emergencystretch=\maxdimen
\usepackage[figuresright]{rotating}

\begin{document}

\begin{frontmatter}



\dochead{}

\title{Adaptive trajectory-constrained exploration strategy for deep reinforcement learning}

\author[label1,label2,label3]{Guojian Wang}
\author[label2,label3,label4,label5]{Faguo Wu \corref{cor1}}
\author[label1,label2,label3,label5]{Xiao Zhang \corref{cor2}}
\author[label1,label2,label3]{Ning Guo}
\author[label2,label3,label4,label5]{Zhiming Zheng}

\address[label1]{School of Mathematical Sciences, Beihang University, Beijing 100191, China}
\address[label2]{Key Laboratory of Mathematics, Informatics and Behavioral Semantics, Ministry of Education, Beijing 100191, China}
\address[label3]{Peng Cheng Laboratory, Shenzhen 518055, Guangdong, China}
\address[label4]{Institute of Artificial Intelligence, Beihang University, Beijing 100191, China}
\address[label5]{Zhongguancun Laboratory, Beijing 100194, China}

\cortext[cor1]{Corresponding author at: Institute of Artificial Intelligence, Beihang University, Beijing 
100191, China. E-mail address: faguo@buaa.edu.cn}
\cortext[cor2]{Corresponding author at: School of Mathematical Sciences, Beihang University, Beijing 
100191, China. E-mail address: xiao.zh@buaa.edu.cn}

\begin{abstract}
  Deep reinforcement learning (DRL) faces significant challenges in addressing the hard-exploration problems in tasks with sparse or deceptive rewards and large state spaces. These challenges severely limit the practical application of DRL. Most previous exploration methods relied on complex architectures to estimate state novelty or introduced sensitive hyperparameters, resulting in instability. To mitigate these issues,  we propose an efficient adaptive trajectory-constrained exploration strategy for DRL. The proposed method guides the policy of the agent away from suboptimal solutions by leveraging incomplete offline demonstrations as references. This approach gradually expands the exploration scope of the agent and strives for optimality in a constrained optimization manner. Additionally, we introduce a novel policy-gradient-based optimization algorithm that utilizes adaptively clipped trajectory-distance rewards for both single- and multi-agent reinforcement learning. We provide a theoretical analysis of our method, including a deduction of the worst-case approximation error bounds, highlighting the validity of our approach for enhancing exploration. To evaluate the effectiveness of the proposed method, we conducted experiments on two large 2D grid world mazes and several MuJoCo tasks. The extensive experimental results demonstrate the significant advantages of our method in achieving temporally extended exploration and avoiding myopic and suboptimal behaviors in both single- and multi-agent settings. Notably, the specific metrics and quantifiable results further support these findings. The code used in the study is available at \url{https://github.com/buaawgj/TACE}.
\end{abstract}

\begin{keyword}
deep reinforcement learning, hard-exploration problem, policy gradient, offline suboptimal demonstrations



\end{keyword}

\end{frontmatter}


\section{Introduction}\label{sec1}

Deep reinforcement learning has achieved considerable success in various fields over the past few years, for example, playing Atari games with raw pixel inputs~\cite{Mnih2015HumanlevelCT,Schulman2015TrustRP}, mastering the game of Go~\cite{Silver2016MasteringTG}, and acquiring complex robotic manipulation and locomotion skills from raw sensory data~\cite{Lillicrap2016ContinuousCW,Schulman2016HighDimensionalCC,fujimoto2018addressing}. Despite these success stories, these DRL algorithms may suffer from poor performance in tasks with sparse and deceptive rewards, and large state spaces\cite{houthooft2016vime,Florensa2017StochasticNN}. We refer to this as a hard-exploration problem, which has received increasing attention~\cite{hong2018diversity,masood2019diversity}. Such hard-exploration tasks are common in the real world. For example, in a navigation task, a reward is only received after the agent collects certain items or reaches terminal points. Our proposed method can be applied to such tasks and help agents explore their environment more systematically.

The hard-exploration problem of reinforcement learning (RL) can be formally defined as an exploration in environments where rewards are sparse and even deceptive~\cite{guo2020memory,gulcehre2019making}. For both single- and multi-agent RL methods, the difficulty lies primarily in the fact that random exploration rarely results in terminated states or meaningful feedback collection. The sparsity of rewards makes the training of neural networks extremely inefficient~\cite{Florensa2017StochasticNN,gangwani2019learning}, because there are no sufficient and immediate rewards as supervisory signals to guide the training of neural networks. The challenge is trickier when troublesome deceptive rewards exist in these tasks because they can lead to myopic behaviors; hence, the agent may often lose the chance to obtain a higher score~\cite{guo2020memory}. Efficient exploration is the key to solving these problems by encouraging the agent to visit underexplored states.

Generally, sample efficiency and premature convergence interfere with the exploration of DRL algorithms in environments with sparse and deceptive rewards and large state spaces~\cite{Florensa2017StochasticNN,guo2020memory,peng2020non,liu2021cooperative}. First, when we use DRL algorithms to solve these hard-exploration tasks in single- and multi-agent settings, such as $\epsilon$-greedy~\cite{Mnih2015HumanlevelCT,sutton2018reinforcement,rashid2020monotonic} or uniform Gaussian exploration noise~\cite{fujimoto2018addressing,schulman2017proximal}, may cause an exponential difference in the sampling complexity for different goals~\cite{osband2016generalization}, which places a high demand on computing power. Second, when an agent frequently collects trajectories for goals with deceptive rewards, it tends to adopt myopic behaviors and learn suboptimal policies. Under the current reinforcement learning paradigm, the agent further limits its exploration to small regions of the state space around suboptimal goals because of the myopic behaviors learned from previous experiences~\cite{peng2020non}. Therefore, the agent will permanently lose the opportunity to achieve a higher score and become stuck in local optima~\cite{guo2020memory}.

In this study, we developed a novel TrAjectory-Constrained Exploration (TACE) method to overcome these challenges. Our method exploits offline suboptimal demonstration data for faster and more efficient exploration without incurring high computational costs and ensuring stability. Our approach orients its policy away from suboptimal policies in the perspective of constrained optimization by considering offline data as a reference. We developed three practical policy-gradient-based algorithms, TCPPO, TCHRL, and TCMAE, with clipped novelty distance rewards. These algorithms can search for a new policy whose state-action visitation distribution is different from the policies represented by offline data. Furthermore, the scale of novelty for a state-action pair is determined based on the maximum mean discrepancy, and this definition of novelty is comparable. Therefore, a distance normalization method was introduced to enable the agent to gradually expand the scope of exploration centered on past trajectories. Thus, the proposed method adaptively adjusts constrained boundaries. To further improve algorithm performance, an adaptive scaling method was developed to ensure that the agent remained inside the feasible region. We then provide a theoretical analysis of our algorithm, deduce the worst-case approximation error bound and theoretically determine the range of hyperparameter values. Finally, sufficient experimental results demonstrated the effectiveness of TACE compared with other state-of-the-art baseline methods for various benchmarking RL tasks.

In summary, our contributions are summarized as follows:
\begin{enumerate}
  \item This paper investigated a trajectory-constrained exploration strategy that promotes sample efficiency and avoids premature convergence for the hard-exploration challenge of single- and multi-agent tasks.
  \item We show a feasible instance where offline suboptimal demonstrations are used as a reference to provide dense and sustainable exploration guidance and enhance the sample efficiency of RL methods.
  \item No additional neural networks are required to model the novelty. Our proposed algorithm is simple in form and explicit in physical meaning, which helps adjust constrained boundaries and expand the exploration scope adaptively. 
  \item A theoretical analysis of our method is provided, explaining the validity of TACE in achieving diverse exploration and the rationality of the TACE design.
  \item The proposed methods were evaluated for various benchmarking tasks, including two large 2D 
  grid world mazes and two MuJoCo mazes. Our method outperforms other advanced baseline methods in terms of exploration efficiency and average returns.
\end{enumerate}

The remainder of this paper is organized as follows: Section~\ref{sec:related} describes the progress of the related work. Section~\ref{sec:backgroung} briefly describes the preliminary knowledge of the article. Section~\ref{sec:approach} introduces the proposed trajectory-constrained exploration strategy. The experimental results are presented in Section~\ref{sec:experience}. Finally, the conclusions are presented in Section~\ref{sec:conclusion}.

\section{Related work}\label{sec:related}
Several methods have been proposed in previous works to encourage sufficient exploration of the agent. Some studies suggest adding noise sampled from a stochastic distribution, such as the Gaussian distribution~\cite{fujimoto2018addressing,schulman2017proximal} or the Ornstein-Unlenbeck process~\cite{Lillicrap2016ContinuousCW}, to actions generated by the policy network, which can motivate the agent to access underexplored areas. Maximum entropy RL methods~\cite{haarnoja2017reinforcement,haarnoja2018soft} allow the agent to explore the environment by encouraging high-entropy distributions over action spaces, given the state inputs. However, such methods are unlikely to achieve satisfactory performance and may result in suboptimal behavior in tasks with sparse and deceptive rewards, long horizons, and large state spaces~\cite{Zhang2019LearningNP}.

At the same time, some methods use intrinsic rewards~\cite{burda2018exploration}, such as surprise~\cite{achiam2017surprise} and curiosity~\cite{stadie2015incentivizing}, to encourage the agent to visit unfamiliar states. For example, curiosity in~\cite{pathak2017curiosity} was formulated as an error in an agent's ability to predict the consequences of its actions in a visual feature space. Although these methods can alleviate the sparse reward problem, they usually rely on auxiliary models to calculate intrinsic rewards, and therefore increase the complexity of the entire model and may incur high computational costs. Another way to cope with sparse or delayed rewards is to define a reward function as a combination of shaping and terminal rewards~\cite{tipaldi2022reinforcement}. However, these methods need to introduce extra hyperparameters to balance the weight of importance between the RL task rewards and intrinsic rewards, which might incur instability.

Diversity-regularized exploration expands an agent's exploration space efficiently, and there have been several recent studies in this area~\cite{hong2018diversity,peng2020non,masood2019diversity}. For instance, collaborative exploration~\cite{peng2020non} employs a team of heterogeneous agents to explore an environment and utilizes a special regularization mechanism to maintain team diversity. Diversity-driven exploration~\cite{hong2018diversity} proposes the addition of a Kullback-Leibler (KL) divergence regularization term to encourage the DRL agent to attempt policies that differ from previously learned policies. Other studies have enabled RL agents to perform exploration more consistently without incurring additional computational costs by adding random noise to the network parameters~\cite{plappert2017parameter,fortunato2019noisy}. However, the diversity term in these methods only considers the divergence in the action space.

DIPG~\cite{masood2019diversity} uses a maximum mean discrepancy (MMD) regularization term to encourage the agent to learn a novel policy that induces a different distribution over the trajectory space. Specifically, the MMD distance was introduced between the trajectory distribution of the current policy and that of the previously learned policies. Moreover, DIPG considers the diversity gradient during training because it adds an $\mathrm{MMD}$ regularization term to its objective function. Therefore, its objective function incurs sensitive hyperparameters that cause instability and even lead to failure to solve the specified task. There are many differences between our proposed method and DIPG, including the definition of the distance measure, formulation of the optimization problem, and design of the optimization solution method. These significant differences are described in detail in Section~\ref{sec:approach}.

Reinforcement Learning from Demonstrations (RLfD) has been proven to be an effective approach for solving problems that require sample efficiency and involve difficult exploration. Demonstrations of RLfD were generated by either the expert or the agent. For example, self-imitation learning~\cite{oh2018self} demonstrates that exploiting previous good experiences can indirectly drive deep exploration. Deep q-learning from demonstrations (DQfD)~\cite{Hester2018DeepQF} leverages even very small amounts of demonstration data to accelerate learning and enhance the exploration of the agent. Recurrent Replay Distributed DQN from Demonstrations (R2D3)~\cite{Paine2020MakingEU} extracts information from expert demonstrations in a manner that guides an agent's autonomous exploration in the environment. The Learning Online with Guidance Offline (LOGO) algorithm~\cite{Rengarajan2022ReinforcementLW} orients the update of its policy by obtaining guidance from offline data, which can significantly reduce the number of exploration actions in sparse reward settings. These methods train the output of the current policies to be close to that of expert policies represented by the demonstration data. However, the cost of obtaining expert demonstration data may be high, and it is often hopeless for the agent to generate highly rewarded trajectories in some hard-exploration tasks. 

HRL has long been recognized as a promising approach to overcoming sparse reward and long-horizon problems~\cite{Dayan1992FeudalRL,sutton1999between,dietterich2000hierarchical,chentanez2004intrinsically}. Recent studies proposed a range of HRL methods to efficiently learn policies in long-horizon tasks with sparse rewards~\cite{levy2019learning,nachum2018data,Florensa2017StochasticNN,heess2016learning}. Under this paradigm, the state-action search space for the agent is exponentially reduced through several modules of abstraction at different levels, and some subsets of these modules may be suitable for reuse~\cite{Florensa2017StochasticNN}. Currently, HRL methods are divided into two main categories. The first category is subgoal-based HRL methods, in which the high-level policy sets a subgoal for the low-level policy to achieve. Distance measurement is required to measure the internal rewards of low-level policies according to current states and subgoals. Some algorithms, such as HAC~\cite{levy2019learning} and HIRO~\cite{nachum2018data}, simply use the Euclidean distance, while FeUdal Networks (FuNs)~\cite{vezhnevets2017feudal} adopt the cosine distance. However, these measurements of state spaces do not necessarily reflect the "true" distance between two states; therefore, these algorithms are sensitive to state-space representation~\cite{dwiel2019hierarchical}.

The second category of HRL methods allows a high-level policy to select a pre-trained low-level skill over several time steps. Therefore, they typically require training in high- and low-level policies for different tasks. Moreover, low-level skills are pre-trained by maximizing diversity objectives~\cite{eysenbach2018diversity}, proxy rewards~\cite{Florensa2017StochasticNN}, or specialized simple tasks~\cite{heess2016learning}. When solving downstream tasks, pre-trained skills are often frozen and only high-level policies are trained, which may lead to significant suboptimality in future tasks~\cite{li2019hierarchical}. The Option-Critic algorithm~\cite{Bacon2017TheOA} makes an effort to train high-level and low-level policies jointly. However, joint training may lead to semantic loss of high-level policies~\cite{vezhnevets2017feudal} and the collapse of low-level skills~\cite{harb2018waiting}.

Although the single-agent exploration problem is extensively studied and has achieved considerable success, few exploration strategies have been developed for multi-agent reinforcement learning (MARL). MAVEN~\cite{mahajan2019maven} encodes a shared latent variable with a hierarchical policy and learns several separate state-action value functions for each agent. EITI~\cite{wang2020influence} uses mutual information (MI) to capture the influence of one agent's behavior on expected returns of the multi-agent team. Previous work~\cite{liu2021cooperative} demonstrates that exploiting structural information on the reward function in MARL tasks can promote exploration. EMC~\cite{zheng2021episodic} introduces a curiosity-driven exploration for episodic MARL by utilizing the results in the episodic memory to regularize the loss function.

\section{Preliminaries}\label{sec:backgroung}
\subsection{Reinforcement Learning}
We consider an infinite-horizon discounted Markov decision process (MDP) defined by a tuple $M = (\mathcal{S}, \mathcal{A}, P, R_e, \rho_{0}, \gamma)$, where $\mathcal{S}$ is a state space, $\mathcal{A}$ is a (discrete or continuous) action space, $P: \mathcal{S} \times \mathcal{A} \times \mathcal{S} \rightarrow \mathbb{R}_{+}$ is the transition probability distribution, $R_e: \mathcal{S} \times \mathcal{A} \rightarrow [R_{min}, R_{max}]$ is the reward function, $\rho_{0}: \mathcal{S} \rightarrow \mathcal{R}_{+}$ is the distribution of the initial state $s_{0}$, and $\gamma \in [0,1]$ is a discount factor. A stochastic policy $ \pi_{\theta}: \mathcal{S} \rightarrow \mathcal{P}(\mathcal{A})$ parametrized by $\theta$, maps the state space $\mathcal{S}$ to a set of probability distributions over the action space $\mathcal{A}$. The standard state-action value function $Q_e$, the value function $V_e$ and the advantage function $A_e$ are defined as follows: 
\begin{equation*}
  Q_e(s_t, a_t) = \mathbb{E}_{s_{t+1}, a_{t+1}, \dots}\left[\sum_{l=0}^{\infty}\gamma^l R_e(s_{t+l}, a_{t+l})\right],
\end{equation*}
\begin{equation*}
  V_e(s_t) = \mathbb{E}_{a_t, s_{t+1}, a_{t+1}, \dots}\left[\sum_{l=0}^{\infty}\gamma^l R_e(s_{t+l}, a_{t+l})\right],
\end{equation*}
\begin{equation*}
  A_e(s, a) = Q_e(s, a) - V_e(s),
\end{equation*}
where $s_t\sim\pi(a_t\vert s_t)$, $s_{t+1}\sim P(s_{t+1}\vert s_t, a_t)$, $\forall t\ge 0$.

Generally, the objective of the RL algorithm is to determine the optimal policy $\pi_{\theta}$ that maximizes the expected discounted return.
\begin{equation*}
J(\pi_{\theta}) = \mathbb{E}_{\tau}\left[\sum_{t=0}^{\infty}\gamma^{t} R_e(s_{t},a_t)\right],
\end{equation*}
here we use $\tau = (s_0, a_0, s_1, a_1, \dots)$ to denote the entire history of state-action pairs in an episode, and $s_0 \sim \rho_0(s_0)$, $a_t \sim \pi_{\theta}(a_t \vert s_t)$, and $s_{t+1} \sim P(s_{t+1}\vert s_t,a_t)$.

State visitation distribution is also of interest. When $\gamma<1$, the discounted state visitation distribution $d^\pi$ is defined as $d^\pi(s) = (1-\gamma)\sum_{t=0}^{\infty}\gamma^t \mathbb{P}(s_t=s\vert \pi, \rho_0)$, where $\mathbb{P}(s_t=s\vert \pi, \rho_0)$ denotes the probability of $s_t=s$ concerning the randomness induced by $\pi$, $P$ and $\rho_0$. 

\subsection{Multi-Agent Reinforcement Learning}
A cooperative multi-agent system can be modeled as a multi-agent Markov decision process. An $n-$agent MDP is defined by a tuple $(\bar{\mathcal{S}}, \bar{\mathcal{A}}, \bar{P}, I, \bar{R}_e, \gamma)$, where $I = \{1, 2, \dots, n\}$ is the finite sets of agents, $\bar{\mathcal{S}} = \times_{i\in I}\mathcal{S}_i$ is the joint state space and $\mathcal{S}_i$ is the state space of $i-$th agent. At each time step, each agent selects an action $a_i \in \mathcal{A}_i$ at state $\mathbf{s}\in\bar{\mathcal{S}}$ to form a joint $\mathbf{a}\in\bar{\mathcal{A}}$, a shared extrinsic reward $\bar{R}_e(\mathbf{s}, \mathbf{a})$ can be received for each agent, and the next state $\mathbf{s}^\prime$ is generated according to the transition function $\bar{P}(\cdot \vert \mathbf{s}, \mathbf{a})$. The objective of the cooperative multi-agent task is that each agent learns a policy $\pi_i(a_i\vert s_i)$ to jointly team performance. In this study, different from this original multi-agent optimization objective, we focus on multi-agent exploration (MAE) and design a novel approach to encourage the agent team to cooperatively explore the environment. 

\subsection{Maximum Mean Discrepancy}
Maximum Mean Discrepancy is an integral probability metric that measures the difference (or similarity) between two different probability distributions~\cite{gretton2006kernel,Gretton2012OptimalKC,thomas2016energetic,masood2019diversity,dziugaite2015training}. Let the probability distributions $p$ and $q$ be defined in a nonempty compact metric space $\mathbb{X}$. Let $x$ and $y$ be observations taken independently of $p$ and $q$, respectively. Subsequently, the MMD metric was defined as
\begin{equation}
  \label{equ:MMD}
  {\rm MMD}(p, q, \mathcal{F}) = \sup_{f \in \mathcal{F}} \left(\mathbb{E}_{x\sim p} \left[f(x)\right] - \mathbb{E}_{y \sim q}\left[f(y)\right]\right),
\end{equation}
where $\mathcal{F}$ is the class of the functions in $\mathbb{X}$. If $\mathcal{F}$ satisfies the condition $p=q$, if and only if $\mathbb{E}_{x\sim p}[f(x)]=\mathbb{E}_{y\sim q}[f(y)],\forall f\in\mathcal{F}$, then MMD is a metric measuring the discrepancy between $p$ and $q$~\cite{fortet1953convergence}.

What type of function class makes up the MMD metrics? According to the literature~\cite{gretton2006kernel}, the space of bounded continuous functions on $\mathbb{X}$ satisfies the condition; however, it is difficult to compute the MMD distance between $p$ and $q$ with finite samples in such a function class because of its uncountability. If $\mathcal{F}$ is the Reproducing Kernel Hilbert Space (RKHS) $\mathcal{H}$, $f$ in Eq.~\eqref{equ:MMD} can be replaced by a kernel function $k\in\mathcal{H}$, such as the Gaussian or Laplace kernel functions. Gretton \emph{et al.}~\cite{Gretton2012OptimalKC} show that
\begin{equation}
  \label{equ:MMD_rkhs}
  \begin{aligned}
    \mathrm{MMD}^2(p, q, \mathcal{H}) = \mathbb{E}[k(x, x^\prime)] - 2\mathbb{E}[k(x, y)] + \mathbb{E}[k(y, y^\prime)],
  \end{aligned}
\end{equation}
where $x, x^\prime\ \mathrm{i.i.d.}\sim p$ and $y, y^\prime\ \mathrm{i.i.d.} \sim q$.
\begin{figure}[htb]
  \centering
  \includegraphics[scale=0.35]{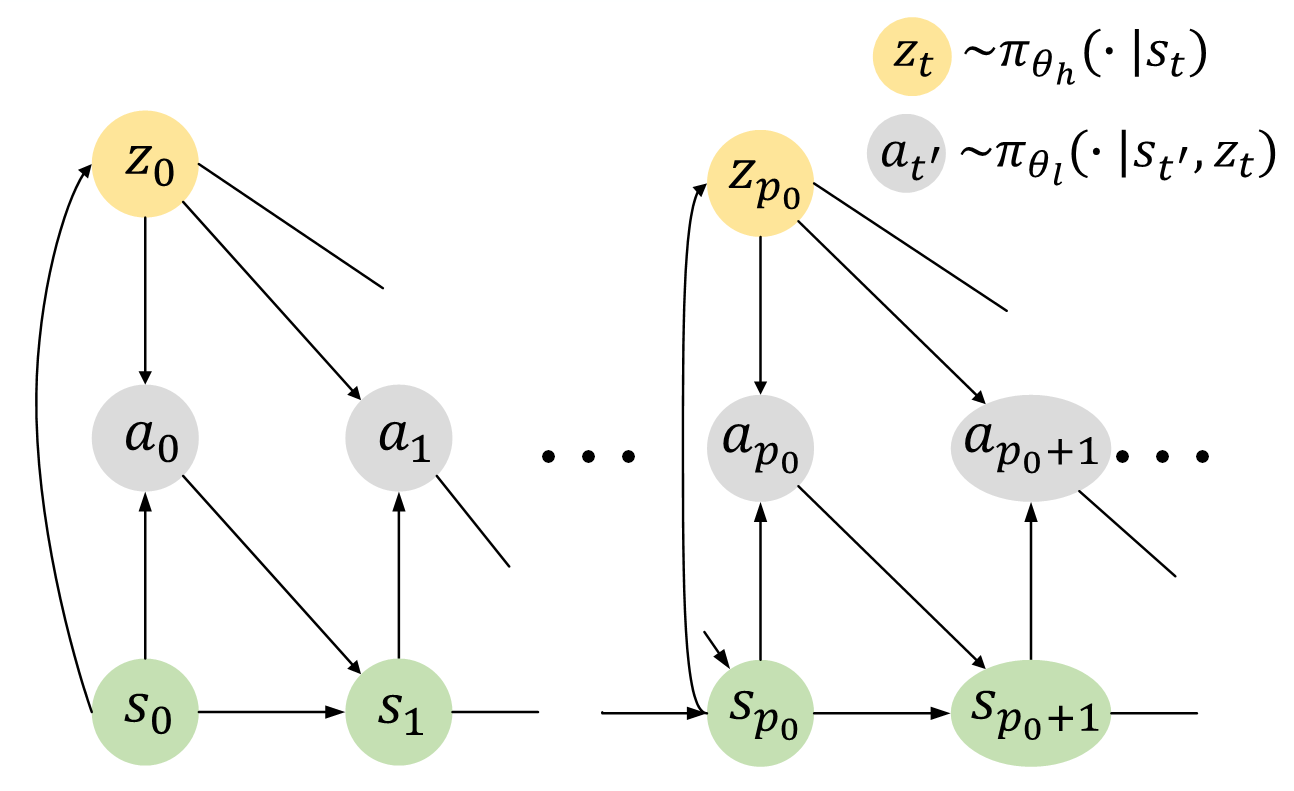}
  \caption{In a two-layer structure, the high-level policy $\pi_{\theta_h}(z_{t} \vert s_{t})$ samples a latent-codes $z_t$, and decides which low-policy performs over next $ p $ time steps. The low-level policy $\pi_{\theta_l}(a_t \vert s_t, z_{t})$ outputs actions $ a_t $ straight interacting with a external environment. Note that $z_{t} = z_{kp}$, from $t=kp$ to $(k+1)p - 1$. After $p$ time steps, the high-level policy takes a new high-level action once again.}
  \label{fig:hierarchy}
\end{figure}

\subsection{Skill-Based HRL Algorithms}
Consider a skill-based HRL algorithm with a 2-level hierarchy, whose policy is composed of a high-level policy $\pi_{\theta_{h}}$ and a low-level policy $\pi_{\theta_{l}}$. In this framework, the skills of a low-level policy are usually pre-trained and distinguished using different latent codes $z$. For example, a single stochastic neural network (SNN)~\cite{tang2013learning} can encode many low-level skills simultaneously~\cite{Florensa2017StochasticNN,li2019hierarchical}. The high-level policy $\pi_{\theta_{h}}(z \vert s)$ does not take actual actions to interact with the external environment directly. Instead, it samples latent codes on a slower timescale than the low-level policy, such as observing the environment once every $p$ time steps. The inputs of low-level policy $\pi_{\theta_{l}}(a \vert s, z)$ are not only environment state vectors but also latent codes from the high-level policy. That is, the high-level policy determines the action output of the low-level policy for the next $p$ time steps~\cite{Bacon2017TheOA, Florensa2017StochasticNN, Li2020SubpolicyAF}. Fig.~\ref{fig:hierarchy} illustrates the framework of an HRL algorithm with a two-level hierarchy.

\section{Proposed Approach}
\label{sec:approach}
In this section, we propose a trajectory-constrained exploration strategy for efficient exploration, which solves the hard-exploration problem from the perspective of constrained optimization. The main objective of the proposed trajectory-constrained exploration strategy is to encourage an RL agent to visit underexplored regions of the state space and prevent it from adopting suboptimal and myopic behaviors.

\subsection{Trajectory-Constrained Exploration Strategy}
\label{sec:optimization_problem}
Assuming that there is a collection $\mathcal{M}$ of offline suboptimal trajectory data that might lead to goals with deceptive rewards, we aim to drive the agent to systematically explore the state space and generate new trajectories by visiting novel regions of the state space. One method to achieve this goal is to maximize the difference between the current and previous trajectories. We used the $\mathrm{MMD}$ metric defined in Eq.~\eqref{equ:MMD} to measure the disparity between the different trajectories. 

The agent collects a certain number of trajectories and stores them in on-policy trajectory buffer $\mathcal{B}$ at each epoch. Specifically, every trajectory $\tau$ in buffer $\mathcal{B}$ and offline replay memory $\mathcal{M}$ is treated as a deterministic policy. Then we calculate their corresponding state-action visitation distributions $\rho_\tau$. Finally, instead of computing the $\mathrm{MMD}$ diversity measurement of the distribution on the trajectory space induced by previous policies~\cite{masood2019diversity}, in our study, the $\mathrm{MMD}$ distance is calculated between different state-action visitation distributions that belong to the offline demonstration trajectory in $\mathcal{M}$ and the current trajectory in $\mathcal{B}$.

Let $K(\cdot, \cdot)$ denote a kernel of the reproducing kernel Hilbert space $\mathcal{H}$, such as a Gaussian kernel, then an unbiased empirical estimate of $\mathrm{MMD}(\tau, \upsilon, \mathcal{H})$ is given by:
\begin{equation*}
  \label{equ:MMD_traj}
  \begin{aligned}
  \mathrm{MMD}^2(\tau, \upsilon, \mathcal{H}) = \underset{x, x^\prime \sim \rho_\tau}{\mathbb{E}} \left[k\left(x, x^\prime\right)\right] - 2\underset{\begin{subarray}{c}x \sim \rho_\tau \\ y \sim \rho_\upsilon\end{subarray}}{\mathbb{E}} \left[k(x, y)\right] + \underset{y, y^\prime \sim \rho_\upsilon}{\mathbb{E}} \left[k(y, y^\prime)\right],
  \end{aligned}
\end{equation*}
where $\tau \in \mathcal{B}$, $\upsilon \in \mathcal{M}$, and $\rho_\tau$ and $\rho_\upsilon$ are the corresponding state-action visitation distributions of $\tau$ and $\upsilon$, respectively. The function $k(\cdot, \cdot)$ is given by:
\begin{equation*}
  k(x, y) = K\left(g(x), g(y)\right).
\end{equation*}
Function $g$ provides the flexibility to adjust the focus of the ${\rm MMD}$ distance metric for different aspects, such as state visits, action choices, or both. In our experiments, we measure the MMD distance only concerning a relevant subset of the information contained in each state-action pair and choose this subset to be the coordinate $c$ of the center of mass (CoM), i.e., the function $g$ maps a state-action pair $(s,a)$ to $c$. Moreover, we believe it should also make sense to let $g(s,a)=(c,a)$, although it may require us to design a new kernel function $K(\cdot,\cdot)$. Finally, we define the distance $D(x, \mathcal{M})$ of the state-action pair $x = (s, a)$ to replay memory $\mathcal{M}$ as follows:
\begin{equation*}
  \label{equ:D}
  D(x, \mathcal{M}) = \underset{\tau\in\mathcal{B}_x}{\mathbb{E}}\left[\mathrm{MMD}^2(\tau, \mathcal{M}, \mathcal{H})\right],
\end{equation*}
where $\mathcal{B}_x = \{\tau \vert x\in\tau, \tau\in\mathcal{B}\}$, and $\mathrm{MMD}^2(\tau, \mathcal{M}, \mathcal{H})$ is defined by:
\begin{equation*}
  \mathrm{MMD}^2(\tau, \mathcal{M}, \mathcal{H}) = \min_{\upsilon\in\mathcal{M}}\mathrm{MMD}^2(\tau, \upsilon, \mathcal{H}).
\end{equation*}

To emphasize that the distance is defined based on the $\mathrm{MMD}$, we add the subscript $\mathrm{MMD}$ to the symbol $D$ in Eq.~\eqref{equ:D}. The stochastic optimization problem with $\mathrm{MMD}$ distance constraints is defined as follows:
\begin{equation}
  \label{equ:constraint}
  \begin{aligned}
    &\max_{\theta} \, J(\theta), \\
    &s.t.\ D_{\rm MMD}(x, \mathcal{M}) \ge \delta, \quad \forall x \in \mathcal{B},
  \end{aligned}
\end{equation}
where $J$ is an ordinary reinforcement learning objective, and $\delta$ is a constant 
{\rm MMD} boundary constraint.

\begin{remark}
  Replay memory $\mathcal{M}$ did not change during any epochs of the training process. Instead, replay memory $\mathcal{M}$ is updated only with trajectories generated by a suboptimal policy learned after a previous training process. Furthermore, the offline trajectories in $\mathcal{M}$ can be collected from human players. The key insight of our study is that it orients its policy away from suboptimal policies by considering incomplete offline demonstrations as references. In contrast to RLfD methods, our method does not require perfect and sufficient demonstrations, which is more realistic in practice. 
\end{remark}

\begin{remark} 
  DIPG~\cite{masood2019diversity} uses the MMD distance between probability distributions over trajectory spaces induced by the current policy and the previously learned suboptimal policy as a diversity regularization term. However, DIPG simply stacks the states and actions from the first $N$ steps of a trajectory into a single vector, which is not sufficient for measuring the distance between trajectory distributions. Instead, we regard each past suboptimal trajectory as a deterministic policy and use the MMD distance metric to calculate the distance between the state-action visitation distributions induced by the current and past suboptimal trajectories. Consequently, our method can reduce the sampling and computational complexities compared with DIPG.
\end{remark}

The constrained optimization problem~\eqref{equ:constraint} can be solved using the Lagrange multiplier method, and its corresponding unconstrained form is as follows: 
\begin{equation}
  \label{equ:unconstrained}
  L(\theta, \sigma) = J(\theta) + \sigma\sum_{x \in \mathcal{B}}\min\left\{D_{\rm MMD}(x, \rho_\mu) - \delta, 0 \right\}.
\end{equation}

The unconstrained form in Eq.~\eqref{equ:unconstrained} is intractable because its second term is numerically unstable, and its gradient concerning the policy parameters is difficult to calculate directly. To address these challenges, we provide a specialized approach to achieve the goal of introducing policy parameters into the second term of the unconstrained optimization problem and balancing the contributions of the RL objective and constraints. According to statistical theory, frequency is an unbiased estimate of probability when the number of samples is sufficiently large. Therefore, when the sample number $N$ of the on-policy buffer $\mathcal{B}$ is sufficiently large, the following formula holds:
\begin{equation*}
  \lim_{N \to \infty}  \frac{1}{N}\sum_{x\in\mathcal{B}}\min\left\{D_{\rm MMD}(x, \rho_\mu)-\delta, 0\right\} = \underset{x\sim\rho_\pi}{\mathbb{E}}\left[\min\left\{D_{\rm MMD}(x, \rho_\mu)-\delta, 0\right\}\right]. 
\end{equation*}
Furthermore, the optimization problem~\eqref{equ:constraint} is converted into an unconstrained form:
\begin{equation}
\label{equ: P_I}
  \begin{aligned}
    L(\theta, \sigma) = J(\theta)
    + \sigma\underset{x\sim\rho_\pi}{\mathbb{E}}\left[\min\left\{D_{\rm MMD}(x, \rho_\mu) - \delta, 0\right\}\right],
  \end{aligned}
\end{equation}
where $\sigma > 0$ is the Lagrange multiplier. Since $N$ is a constant, it is absorbed by choosing the appropriate coefficient $\sigma$.

Second, we estimate the gradient of the unconstrained optimization problem concerning the policy parameters. The first term of the unconstrained problem is an ordinary RL objective, and hence, the gradient of this term can be calculated easily~\cite{schulman2017proximal,Schulman2015TrustRP,Mnih2015HumanlevelCT}. We then derive the gradient of the ${\rm MMD}$ term in Eq.~\eqref{equ: P_I} for policy parameters $\theta$, which enables us to efficiently optimize the policy. The result is described in the following lemma.

\begin{restatable}[Gradient Derivation of the MMD term]{lemma}{mmdgradient}
  \label{lemma:1}
  Let $\rho_{\pi}(s,a)$ be the state-action visitation distribution induced by the current policy $\pi$. Let $D(x,\mathcal{M})$ be the MMD distance between the state-action pair $x$ and replay memory $\mathcal{M}$. Then, if the policy $\pi$ is parameterized by $\theta$, the gradient of the ${\rm MMD}$ term of Eq.~\eqref{equ: P_I} for parameters $\theta$ is derived as follows:
  \begin{equation}
    \label{equ:nabla_E_D_mmd}
    \begin{aligned}
      \nabla_{\theta} D_{\rm MMD}=\underset{\rho_{\pi}(s,a)}{\mathbb{E}}
      \left[\nabla_{\theta}\log\pi_{\theta}(a \vert s)Q_i(s,a)\right],
    \end{aligned}
  \end{equation}
  where
  \begin{equation}
    \label{equ:Q}
    Q_i(s_t, a_t) = \underset{\rho_{\pi}(s, a)}{\mathbb{E}}\left[\sum_{l=0}^{T-t}\gamma^{l}R_i(s, a)\right],
  \end{equation}
  and
  \begin{equation}
    R_i(s, a) = \min\left\{D_{\rm MMD}(x, \mathcal{M})-\delta, 0\right\}.
  \end{equation}
\end{restatable}

~\ref{sec:app_3} presents the derivation of the MMD term gradient gradient. 

\begin{remark}
  In Algorithm~\ref{algo:tcppo}, we showed that this trajectory-constrained exploration strategy can be readily applied to on-policy algorithms, such as PPO~\cite{schulman2017proximal}. Whereas the PPO algorithm does not maintain replay memory, our method maintains $n$ trajectories of each past suboptimal policy in replay memory $\mathcal{M}$ to compute the $D_{\rm MMD}$ distance measure. Generally, $n=5$ is sufficient to achieve satisfactory performance. The batch of on-policy data is stored in buffer $\mathcal{B}$. Moreover, because the MMD distance term is treated as a diversity reward in Lemma~\ref{lemma:1}, we believe that it can also be integrated into bootstrapped Q-learning by adding the MMD diversity bonus to both the immediate reward and the next $Q$ value, such as OB2I~\cite{bai2021principled}, which deserves further investigation.
\end{remark}

\begin{remark} 
  DIPG~\cite{masood2019diversity} considers the diversity gradient during the training process because it adds an $\mathrm{MMD}$ regularization term to its objective function. Therefore, its objective function incurs sensitive hyperparameters that cause instability and even lead to failure to solve the specified task. To resolve the contradiction between the RL and diversity objectives, we reformulate the hard-exploration task as a constrained optimization problem, where the optimization goal is formulated by the naive RL objective, and the $\mathrm{MMD}$ constraints bind the agent's exploration region away from the offline demonstrations above a certain threshold. Using this formulation, the offline trajectories regulate the policy updating only when the current trajectories are outside the constraint region, which ensures stability and enables the agent to eliminate the local optima.
\end{remark}

\subsection{Fast Adaptation Methods}
\subsubsection{Adaptive Constraint Boundary Adjustment Method}
Generally, the constraint boundary $\delta$ in Eq.~\eqref{equ:constraint} should be environmentally dependent for different tasks, which remains a great challenge for us to determine the size of parameter $\delta$ in various tasks. Moreover, the expected MMD distance of each epoch gradually increases as the training process continues because of the gradient of the MMD term derived in Lemma~\ref{lemma:1}. Consequently, it is not conducive for the agent to achieve temporally extended exploration when adopting a constant distance constraint boundary $\delta$ during the entire training process. To resolve the problem of an increased MMD distance, we propose an adaptive constraint distance normalization method, inspired by the batch normalization method~\cite{ioffe2015batch}. After the agent generates a batch of $N$ trajectories $\{\tau_i\}_{i}^{n}$ and stores them in $\mathcal{B}$, the distance $d(x, \mathcal{M}) = D_{\rm MMD}(x,\mathcal{M})$ of each state-action pair $x$ in the on-policy buffer $\mathcal{B}$ is calculated in each epoch. We then compute the normalized distance $\hat{d}(x, \mathcal{M})$ according to:
\begin{equation}
  \label{equ:normal}
  \hat{d}(x, \mathcal{M}) = \frac{d(x, \mathcal{M}) - \mathbb{E}\left[D(\mathcal{B})\right]}{\sqrt{{\rm Var}\left[D(\mathcal{B})\right]}},
\end{equation}
where the expectation $\mathbb{E}[D(\mathcal{B})]$ and variance ${\rm Var}[D(\mathcal{B})]$ are computed over the distance 
set $D(\mathcal{B})=\left\{d(x_i, \mathcal{M})\vert x_i\in\mathcal{B})\right\}$ of all state-action pairs in $\mathcal{B}$ in every epoch.

Intuitively, our proposed policy gradient updates the policy parameters to drive the agent to visit the underexplored state-action pairs while increasing the return value of the agent. Hence the expected MMD distance of state-action pairs in $\mathcal{B}$ for each epoch increases gradually during the training process. Distance normalization enables us to adjust distance constraint boundaries dynamically because true distance boundaries represented by the normalized parameter $\delta$ adaptively change for each epoch. With distance normalization, the problem of determining parameter $\delta$ becomes relaxed and environmentally independent. In our experiments, we usually choose $\delta=0.5$ as the distance constraint boundary. We compare the performance of our algorithm with different parameter choices in Section~\ref{subsec:two-goals}. The experimental results demonstrate that this method can stabilize the training process and enable the agent to achieve better temporally extended exploration.

\subsubsection{Adaptive Scaling Method}
\label{subsec:asm}
Although the value of the Lagrange multiplier $\sigma$ can be updated by gradient ascent, we find that this method is less than ideal in some cases, as demonstrated by the experimental results presented in Section~\ref{sec:experience}. To solve this problem, similar to~\cite{hong2018diversity} and~\cite{plappert2017parameter}, we propose an adaptive scaling method (ASM) based on the MMD distance to adjust the contributions of the ordinary RL objective and the MMD term to the gradient. We associate $\sigma$ with the MMD distance metric $D_{\rm MMD}$ and adaptively increase or decrease the value of $\sigma$ depending on whether the MMD distance $D_{\rm MMD}$ between current trajectories and previous suboptimal trajectories are below or above a certain distance threshold $\epsilon$. Different values of the threshold $\epsilon$ are adopted by different methods in our experiments. The simple approach employed to update $\sigma$ for each training iteration~\cite{hong2018diversity,plappert2017parameter} is given by:
\begin{equation}
  \sigma = 
  \begin{cases}
    1.05\sigma, &{\rm if} \,\exists\upsilon\in\mathcal{B}\,\, s.t.\, {\rm MMD}(\upsilon, \mathcal{M}) \le \epsilon, \\ 
    0.98\sigma, &{\rm if} \,\forall\upsilon\in\mathcal{B}\,\, s.t.\, {\rm MMD}(\upsilon, \mathcal{M}) \ge 2\epsilon,
  \end{cases}
  \label{equ:asm}
\end{equation}
where ${\rm MMD}(\upsilon, \mathcal{M}):=\min_{\tau\in\mathcal{M}}{\rm MMD}(\upsilon, \tau)$.
In addition, if trajectories generated by the current policy lead to the same sparse 
reward as previously stored trajectories, then we adjust the value of $\sigma$ to 1.2 times the current value, i.e. $\sigma = 1.2\sigma$. It is worth noting that the values of 1.05, 0.995, and 1.2 are selected empirically. However, they have a certain universality, and we use these values in all the experiments of this study.

\section{Theoretical Analysis of TACE Performance Bounds}
In this section, we present the theoretical foundation of TACE and demonstrate how it improves the exploratory performance of the agent. We derive a novel bound for the difference in returns between two arbitrary policies under the proposed MMD constraints. This result can be viewed as an extension of the previous work~\cite{kakade2002approximately,pirotta2013safe,Schulman2015TrustRP,achiam2017constrained} on the new constrained policy search problem in Eq.~\eqref{equ:constraint}. Against this background, this section provides some guarantees of performance improvement and helps us better understand our proposed algorithms at a theoretical level.

The $D_{\rm MMD}$ gradient estimation in Eq.~\eqref{equ:nabla_E_D_mmd} in Lemma~\ref{lemma:1} is similar to policy gradients introduced in~\cite{schulman2017proximal}, except that environmental rewards $R_e(s, a)$ are replaced by MMD-based diversity constraints $\min\left\{D_{\rm MMD}(x, \mathcal{M})-\delta, 0\right\}$. Thus, TACE considers the long-term effects of constraints while first guaranteeing that the constraints are valid. When the MMD gradient of Lemma~\ref{lemma:1} is integrated with the gradient of $J(\theta)$ to update the policy parameters, the final gradient $g_\theta$ for the parameter update is expressed as
\begin{equation}
  g_\theta = \mathbb{E}_{\rho_\pi(s,a)}\left[\nabla_{\theta}\log\pi_{\theta}(a \vert s)\left(Q_e(s, a) + \sigma Q_i(s,a)\right)\right].
\end{equation}

Due to the similarity between the forms of the $D_{\rm MMD}$ gradient and the RL gradient of $J(\theta)$, MMD-based diversity constraints $\min\left\{D_{\rm MMD}(x, \mathcal{M})-\delta, 0\right\}$ can be viewed as an intrinsic reward $R_i(s, a)$ for each state-action pair and be integrated with environmental rewards as follows:
\begin{equation*}
  Q_e(s_t, a_t) + \sigma Q_i(s_t,a_t) = \mathbb{E}_{s_{t+1}, a_{t+1}, \dots}\left[\sum_{l=0}^{\infty}\gamma^l \left(R_e(s_{t+l}, a_{t+l}) + \sigma R_i(s_{t+l}, a_{t+l})\right)\right],
\end{equation*}
where $R_e(s_{t+l}, a_{t+l}) + \sigma R_i(s_{t+l}, a_{t+l})$ can be viewed as the total reward $R(s,a)$. The following theorem connects the difference in returns between two arbitrary policies to their average variational divergence. 
\begin{restatable}{theorem}{mainthm}[Performance bound of the trajectory-constrained exploration strategy]
  \label{the:bound}
  For the proposed constrained optimization problem of Eq.~\eqref{equ:constraint}. Subsequently, for any policies $\pi$ and $\pi^\prime$, define $\delta_f (s,a,s') \doteq R_e(s, a) + \sigma R_i(s_t, a_t) + \gamma f(s') - f(s)$,
  %
  \begin{equation*}
    \epsilon_f^{\pi^\prime} \doteq \max_s \left\vert \mathbb{E}_{a \sim \pi^\prime} [\delta_f (s,a, s^\prime)] \right\vert,
  \end{equation*}
  \begin{equation*}
    \label{surrogate_0}
    T_{\pi,f} (\pi') \doteq \underset{\begin{subarray}{c} s\sim d^{\pi} \\ a \sim \pi \end{subarray}}{\mathbb{E}} \left[ \left(\frac{\pi'(a|s)}{\pi(a|s)} - 1 \right) \delta_f(s,a, s^\prime) \right], \text{ and }
  \end{equation*}
  \begin{equation*}
    D_{\pi,f}^{\pm} (\pi') \doteq  \frac{T_{\pi,f} (\pi')}{1-\gamma} \pm \frac{2\gamma \epsilon_f^{\pi'}}{(1-\gamma)^2} \underset{s \sim d^{\pi}}{\mathbb{E}} \left[ D_{TV} (\pi'||\pi)[s] \right],
  \end{equation*}
  where $s^\prime\sim P(\cdot\vert s,a)$, $R_i(s, a)$ is the MMD-based distance reward $\min\left\{D_{\rm MMD}(x, \mathcal{M})-\delta, 0\right\}$ and $\gamma$ is the discount factor. $D_{TV}(\pi'||\pi)[s] = \frac{1}{2}\sum_a \left| \pi'(a|s) - \pi(a|s) \right|$ is used to represent the total variational divergence between two action distributions of $\pi$ and $\pi^\prime$ when the state is $s$. The following bounds hold:
  \begin{equation}
    D_{\pi,f}^{+} (\pi') \geq L(\theta^\prime, \sigma) - L(\theta, \sigma) \geq D_{\pi,f}^{-} (\pi'), 
    \label{equ:bound1}
  \end{equation} 
  Furthermore, the bounds are tight (when $\pi' = \pi$, all three expressions are identically zero). Here, $L(\cdot, \cdot)$ is defined in Eq.~\eqref{equ:unconstrained}, $\sigma$ is the Lagrange multiplier.
\end{restatable}

Before proceeding, it is worth noting that Theorem~\ref{the:bound} is similar to Theorem 1 of~\cite{achiam2017constrained}. When choosing $\sigma=0$, the result of this theorem degenerates into Theorem 1 of~\cite{achiam2017constrained}. According to Lemma~\ref{lemma:1}, our approach transforms the constrained optimization problem into an unconstrained policy search task. In this manner, it considers the long-term effects of constraints on returns. Moreover, different from Theorem 1 in~\cite{achiam2017constrained}, our method derives a new performance bound $D_{\pi,f}^{\pm}(\pi^\prime)$ based on $\delta_f (s,a,s') \doteq R_e(s, a) + \sigma R_i(s_t, a_t) + \gamma f(s') - f(s)$. Hence, this theorem can be used to analyze the effectiveness of our approach in improving exploration. By bounding the expectation of the total variational divergence ${\mathbb{E}}_{s\sim d^\pi} \left[ D_{TV} (\pi'||\pi)[s] \right]$ with $\max_{s} \left[ D_{TV} (\pi'||\pi)[s] \right]$, picking $f(s)$ to be the value function $V(s)$ computed with the total reward $R(s,a)$, the following corollary holds:
\begin{restatable}{corollary}{advbound}
  \label{cor:1}
  For any policies $\pi$ and $\pi^\prime$, with $\epsilon^{\pi'} \doteq \max_s | \mathbb{E}_{a \sim \pi'} [A(s,a)] |$, in which $A(s,a)$ is the advantage function calculated with the total reward $R(s, a) \doteq R_e(s, a) + \sigma R_i(s_t, a_t)$, then the following bound holds: 
  \begin{equation}
    \label{equ:bound2} 
    \begin{aligned}
      L(\theta^\prime, \sigma) - L(\theta, \sigma)
      \ge \frac{1}{1-\gamma} \underset{\begin{subarray}{c} s \sim d^{\pi} \\ a \sim \pi' \end{subarray}}{\mathbb{E}} \left[A(s,a) - \frac{2\gamma \epsilon^{\pi'}}{1-\gamma}  D_{TV} (\pi'||\pi)[s] \right].
    \end{aligned}
  \end{equation}
  Here, $L(\cdot, \cdot)$ is defined in Eq.~\eqref{equ:unconstrained}, $\gamma$ is the discount factor, and $\sigma$ is the Lagrange multiplier.
\end{restatable}

The bound in Corollary~\ref{cor:1} can be regarded as the worst-case approximation error. The TV-divergence and KL-divergence are related by $D_{TV} (p||q) \le \sqrt{D_{KL} (p||q) /2}$~\cite{csiszar2011information}. Combining this inequality with Jensen’s inequality, we obtain:
\begin{equation}
  \underset{s \sim d^{\pi}}{\mathbb{E}}\left[D_{TV}(\pi'||\pi)[s]\right] \le \underset{s \sim d^{\pi}}{\mathbb{E}}\left[\sqrt{\frac{1}{2} D_{KL}(\pi'||\pi)[s]}\right]\le \sqrt{\frac{1}{2}\underset{s\sim d^{\pi}}{\mathbb{E}}\left[D_{KL}(\pi'||\pi)[s]\right]}.
  \label{equ:tvkl}
\end{equation}

It is worth mentioning that the advantage $A(s,a)$ can be decomposed as the sum of the environmental advantage $A_e (s,a)$ and the MMD-based advantage $A_{i} (s,a)$, which is expressed as:
\begin{equation}
  \label{equ:A_decomposed}
  A(s,a) = A_e (s,a) + A_{i} (s,a).
\end{equation} 

Substituting Eq.~\eqref{equ:tvkl} and~\eqref{equ:A_decomposed} into Eq.~\eqref{equ:bound2}, we obtain the following corollary about determining the value of $\sigma$, such that the worst-case approximation error in Eq.~\eqref{equ:bound2} is greater than the threshold $\Delta$:
\begin{restatable}{corollary}{findsigma}
  \label{cor:2}
  Suppose a performance improvement threshold $\Delta$ for any policies $\pi$ and $\pi^\prime$, and $\pi$ and $\pi^\prime$ satisfy $\mathbb{E}_{s \sim d^{\pi}}\left[D_{KL}(\pi'||\pi)[s]\right] \le \eta $ and $\mathbb{E}\left[A_{i} (s,a)\right] \ge \beta > 0$. When $(1-\gamma)\Delta  - \mathbb{E}\left[A_e (s,a)\right] - {\sqrt{2\eta} \gamma \epsilon^{\pi^\prime}}(1 -\gamma)^{-1} > 0$, then if 
  \begin{equation}
    \label{eq:sigma}
    \sigma \ge {\underset{\begin{subarray}{c} s \sim d^{\pi} \\ a \sim \pi' \end{subarray}}{\mathbb{E}}^{-1}\left[A_{i} (s,a)\right]} \left[(1-\gamma)\Delta  - \underset{\begin{subarray}{c} s \sim d^{\pi} \\ a \sim \pi' \end{subarray}}{\mathbb{E}}\left[A_e (s,a)\right] - \frac{\sqrt{2\eta} \gamma \epsilon^{\pi^\prime}}{1 -\gamma}\right],
  \end{equation}
  we have
  \begin{equation}
    L(\theta^\prime, \sigma) - L(\theta, \sigma) \ge \Delta.
  \end{equation}
  Here, $L(\cdot, \cdot)$ is defined in Eq.~\eqref{equ:unconstrained}, $\gamma$ is the discount factor, and $\sigma$ is the Lagrange multiplier.
\end{restatable}

This corollary illustrates the feasibility of the adaptive scaling method in Section~\ref{subsec:asm}. According to Corollary~\ref{cor:2}, by choosing suitable parameters $\sigma$ and $\delta$, we can make $L(\theta^\prime, \sigma) - L(\theta, \sigma) \ge \Delta$ hold. Note that when ${\mathbb{E}}\left[A_e (s,a)\right]$ decreases, for example, if the agent adopts a single behavioral pattern and learns a suboptimal policy, a larger lower bound of $\sigma$ is calculated. In this manner, our approach helps the agent to be exempt from myopic behaviors and drives it to stay in the feasible region. In practice, even if we use the minimal value of $\sigma$ recommended by the corollary above, $\sigma$ is still very large to obtain the performance improvement of $\Delta$. Moreover, $\sigma$ is only computed when the parameters of the policy have been updated. Hence, determining the value of $\sigma$ before the update of the policy parameters in each iteration is difficult. One way to take a smaller value of $\sigma$ robustly is to use the heuristic adaptive scaling method.

\section{Experimental Setup}
\subsection{Environments}
\begin{figure}[htb]
  \centering
  \subfloat[]{
    \label{fig:discrete_maze}
    \includegraphics[scale=0.3]{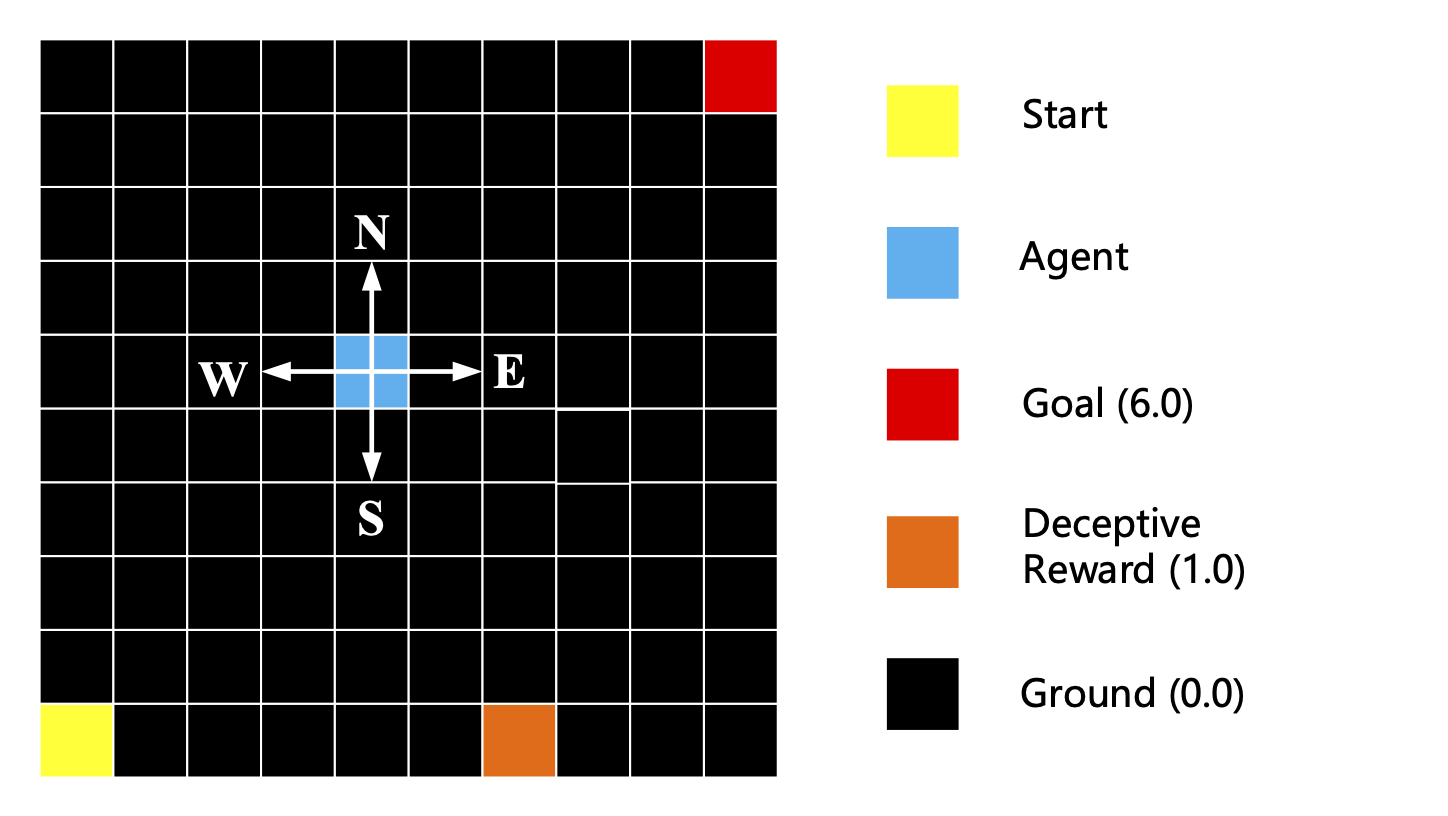}
  }
  \subfloat[]{
    \centering
    \label{fig:deceptive_reacher}
    \includegraphics[scale=0.21]{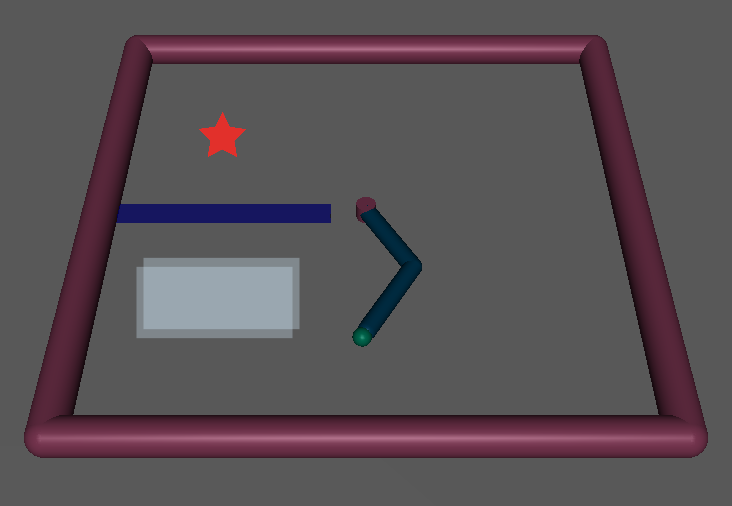}
  }
  \caption{(a) Gridworld; (b) Deceptive Reacher.}
  \label{fig:grid+reacher}
\end{figure}

\noindent\textbf{Gridworld.} We evaluated the performance of the TCPPO (PPO with TACE) algorithm in tasks with discrete state and action spaces, as shown in Fig.~\ref{fig:grid+reacher}\subref{fig:discrete_maze}. In this experimental setting, the agent started from the bottom-left corner of the map, and the optimal goal with the highest reward of 6 is located in the top-right corner. Moreover, there is a suboptimal goal with a relatively small reward of 1 on the right side of the initial position that can be accessed by the agent more easily. The deceptive reward provided by a suboptimal goal can easily distract the agent from finding the goal with the highest reward in the top-right corner. At each time step, the agent observes its coordinates relative to the starting point and chooses from four possible actions: \emph{moving east, south, west,} and \emph{north}. An episode terminates immediately once the agent reaches either of the two goals or the maximum number of steps for an episode is exceeded.

\noindent\textbf{Deceptive Reacher.}
To test the proposed TCPPO method in continuous robotic settings, we used a variant of the classic 3D-Reacher environment with formulable obstacles and misleading rewards~\cite{Zhang2019LearningNP}. In this problem, as shown in Fig.~\ref{fig:grid+reacher}\subref{fig:deceptive_reacher}, a two-joint robot arm~\cite{todorov2012mujoco} attempts to move its end effector (fingertip) close to the red target position to obtain an optimal reward of 60. Instead, the end effector obtains a small deceptive reward of 10 more easily by entering the box. At the start of a new episode, the robot arm is spawned at a random position sampled from a specific range. The agent's observation space consisted of the angles and angular velocities of the two arms and the coordinates of the reacher's fingertips. Furthermore, the actions performed by the agent are sampled from a two-dimensional continuous action space~\cite{1606.01540}.

\begin{figure}[htb]
  \centering
  \subfloat[]{
    \includegraphics[width=5.5cm]{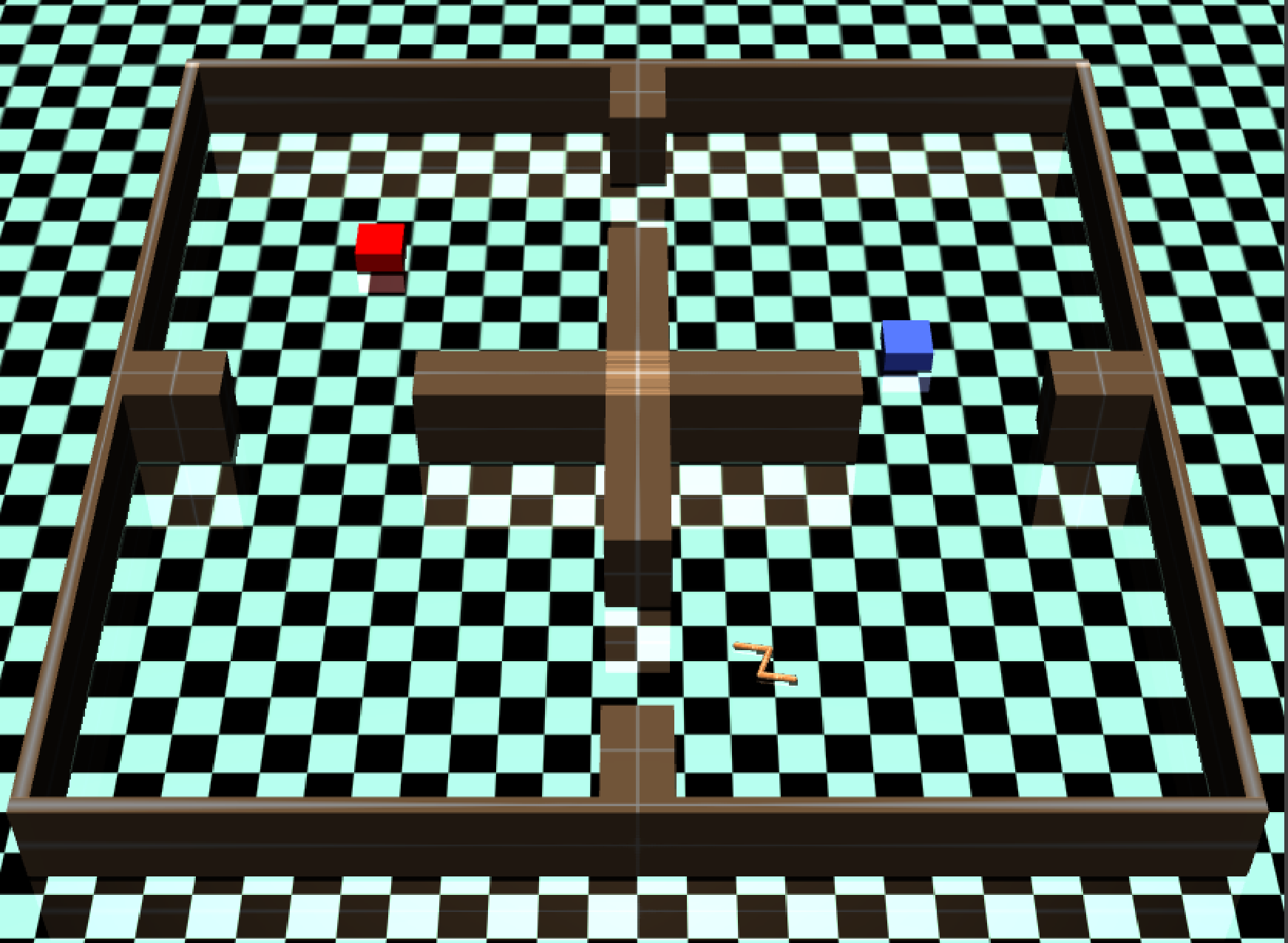}
    \label{fig:maze_0}
  }
  \quad\quad\quad
  \subfloat[]{
    \includegraphics[width=5.5cm]{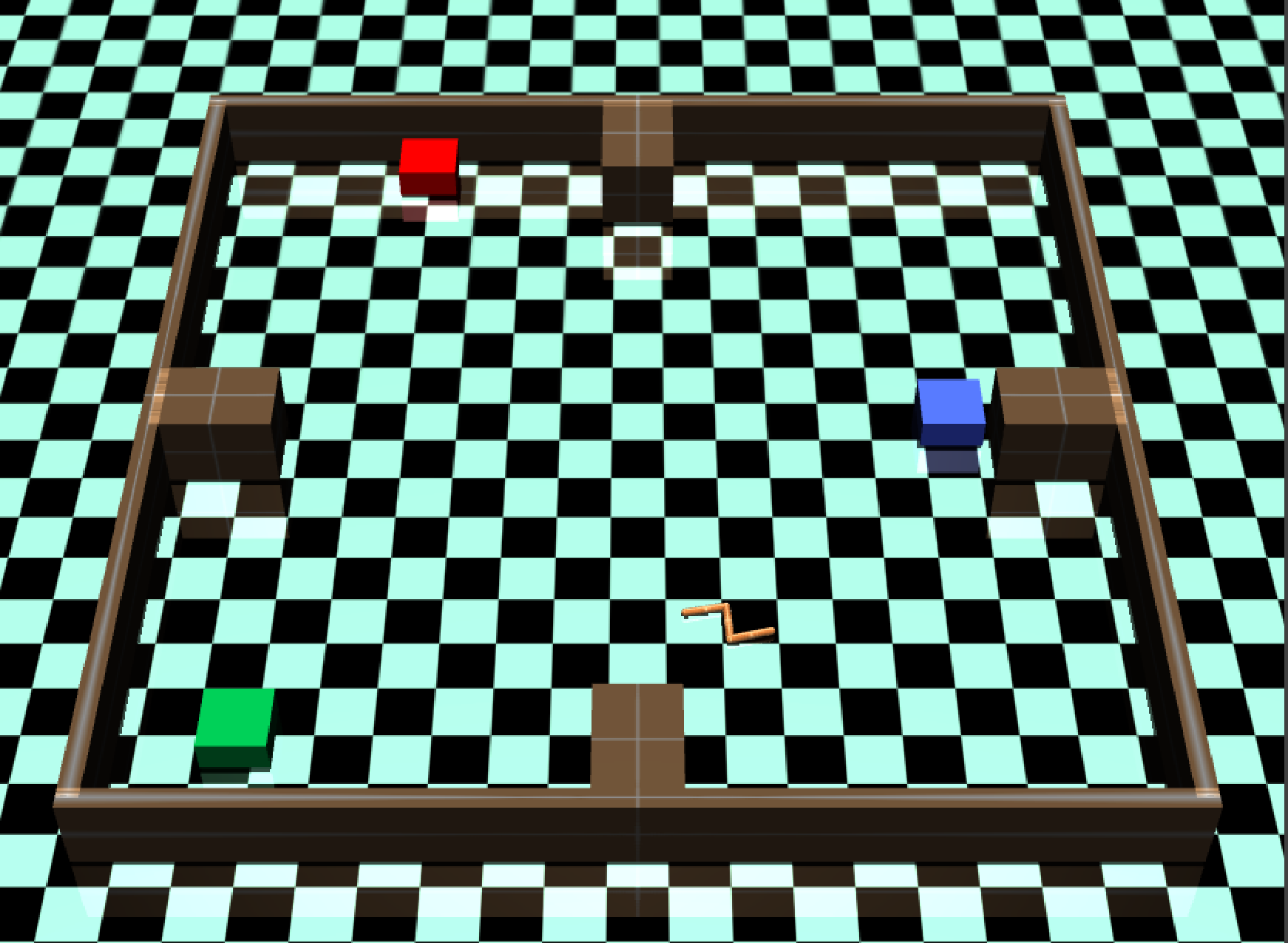}
    \label{fig:maze_1}
  }
  \caption{
    (a) Maze 0 with two different goals. The agent is rewarded 60 for reaching the red goal and 50 for reaching the blue goal. 
    (b) Maze 1 with three different goals. The agent is rewarded 90 for reaching the red goal, 60 for reaching the green goal, and 30 for reaching the blue 
    goal.
  }
  \label{fig:mazes}
\end{figure}

\noindent\textbf{Hierarchical Control Tasks.} 
Two MuJoCo mazes with continuous state-action spaces were adapted from the benchmarking hierarchical tasks used in~\cite{Florensa2017StochasticNN} and~\cite{levy2019learning}, which were used to test TCHRL (SNN4HRL with TACE). The observation space of the agent in these tasks is composed of the internal information $S_a$ of the agent, such as the agent's joint angles, and the task-specific characteristics $S_e$, for example, walls, goals, and other objects seen through a range sensor. These robots were described in~\cite{duan2016benchmarking}. In these tasks, the agent is rewarded for reaching a specified position in the maze, as shown in Fig.~\ref{fig:mazes}. The problem of sparse rewards and long horizons continues to pose significant challenges to RL because an agent rarely obtains nonzero rewards; therefore, the gradient-based optimization of RL for parameterized policies is incremental and slow.

As shown in Fig.~\ref{fig:mazes}\subref{fig:maze_0}, the structure of Maze 0 is the same as that introduced in~\cite{levy2019learning}, except that our maze is larger in scale and has two goals placed in the upper-left and upper-right rooms. The agent can receive rewards of 60 for reaching the goal in the top-left room and 50 for reaching another goal. The agent was initially positioned in the bottom-right room of this maze.

Compared to Maze 0, Maze 1 has a different structure and more goals. Maze 1, shown in Fig.~\ref{fig:mazes}\subref{fig:maze_1}, has three different goals, located at the top-left corner, bottom-left corner, and right side of the maze. The agent obtains rewards of 90, 60, and 30 to reach each goal, and its initial position is near the bottom-left corner. The agent can be more easily distracted from finding the optimal goal using two suboptimal goals. 

\noindent\textbf{Multi-Agent Control Tasks.} 
TCMAE (Multi-Agent Exploration with TACE) was evaluated on two challenging environments: (1) a discrete version of the multiple-particle environment (MPE)~\cite{lowe2017multi,wang2020influence}; and (2) a modified MuJoCo continuous control task - SparseAnt Maze. In both environments, sparse rewards are only collected when agents reach the specified locations. Moreover, agents can receive deceptive suboptimal rewards from targets that are closer to them. 
\begin{figure}[htb] 
  \centering
  \subfloat[]{
    \includegraphics[width=8.5cm]{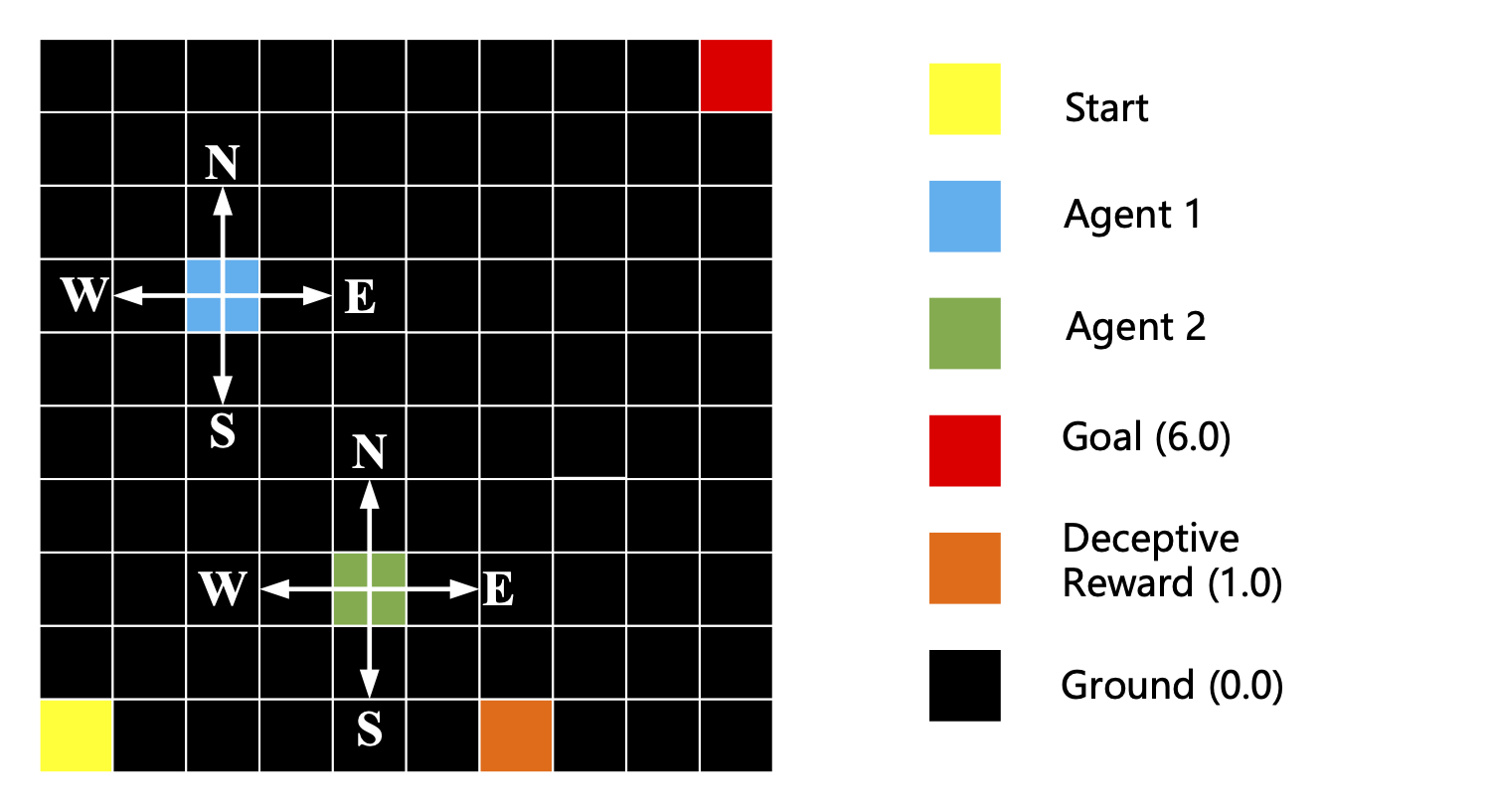}
    \label{fig:ma_grid}
  }
  \subfloat[]{
    \includegraphics[width=5cm]{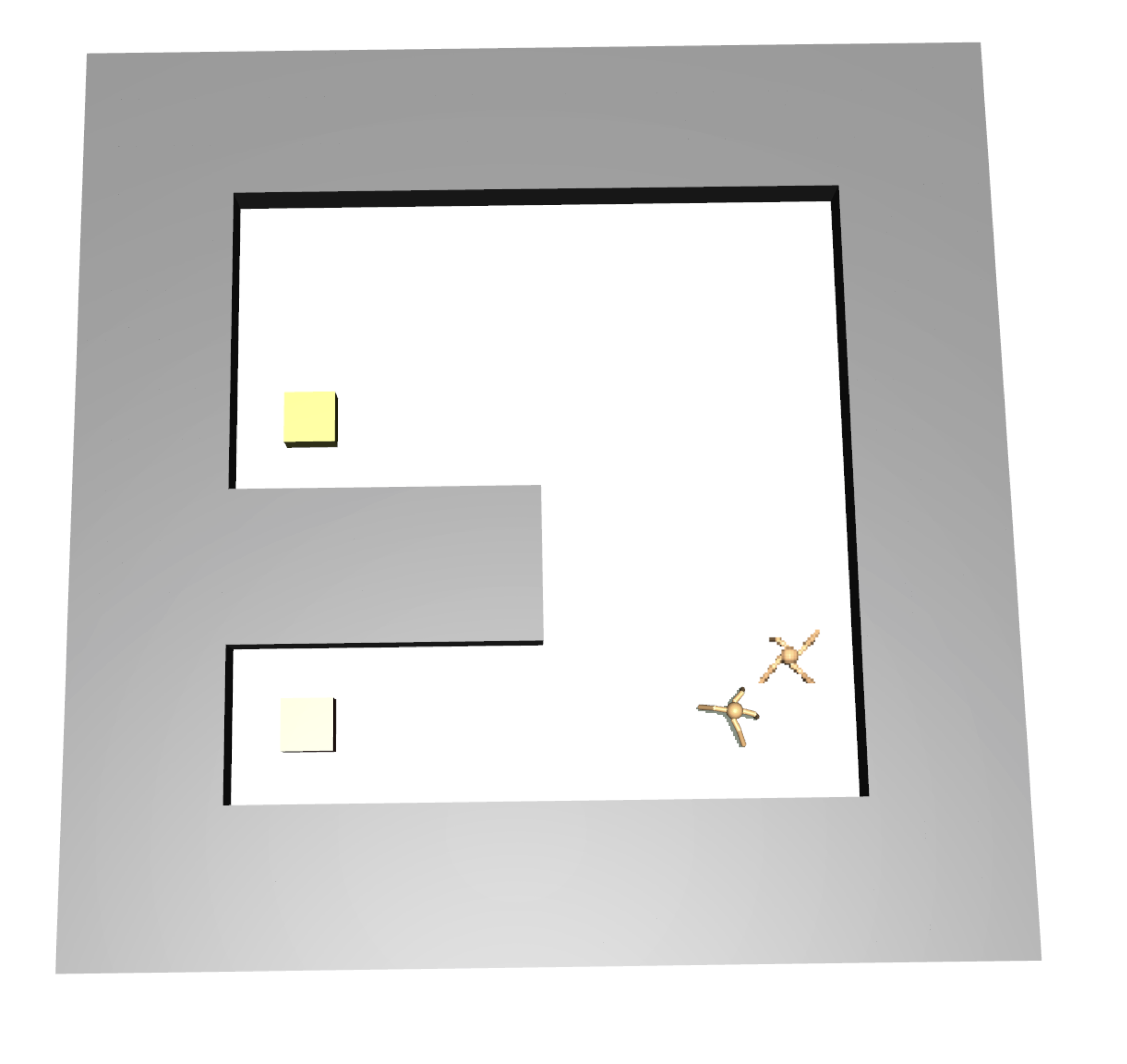}
    \label{fig:ma_ant}
  }
  \caption{
    (a) Discrete multiple-particle environment
    (b) SparseAnt Maze
  }
  \label{fig:ma_tasks}
\end{figure}

As shown in Fig.~\ref{fig:ma_tasks}\subref{fig:ma_grid}, two agents operate within the room of a $70\times 70$ grid in the MPE environment. There is a suboptimal goal of a deceptive reward near the initial position, and the optimal goal is located in the up-right corner of the grid. The agents in the team can receive reward signals only when they reach any goal in the environment. The action space of each agent consists of four discrete actions: \emph{moving east, south, west,} and \emph{north}, similar to the grid-world environment. The agent team hopes to get rid of the suboptimal goal and cooperatively collect the optimal reward by sharing the trajectory information. 

In Fig.~\ref{fig:ma_tasks}\subref{fig:ma_ant}, two MuJoCo Ant agents are rewarded for reaching the specified position in a maze, and other than that, they cannot receive any reward signal. The ant agents start at the same location and cooperatively explore the environment similar to the agents in the MPE task. The observation space of the ant agent is composed of the internal information $S_a$ of the agent and the task-specific characteristics $S_e$ as described in the hierarchical control tasks.

\subsection{Neural Architectures and Hyperparameters}
For grid world tasks, all policies were trained with a learning rate of 0.000018 and a discount factor of 0.99. All neural networks of the policies were represented with fully connected networks that have two layers of 64 hidden units. The batch size and maximum episode length in the $50\times 50$ grid world maze were 8 episodes and 160, respectively. Meanwhile, The batch size was also 8 episodes in the $70\times 70$ grid world maze, while the maximum episode length was 220. The initial value of $\sigma$ is set to 0.5 for these two tasks.

For the deceptive reacher task, all policies were trained with a learning rate of 0.000006 and a discount factor of 0.99. We used fully connected networks that have two layers of 64 hidden units to implement the agent policy. The batch size and maximum episode length were 8 episodes and 150 steps, respectively. The initial value of $\sigma$ is set to 0.5 for TCPPO.

For the MuJoCo maze tasks, we used an SNN to learn pre-trained skills that were trained by TRPO. The number of pre-trained skills (i.e., the dimensions of the latent code fed into the SNN) was six. A high-level policy is implemented using a fully connected neural network trained by the PPO. All the neural networks (SNN and fully connected networks) had two layers of 64 hidden units. The other settings were identical to those in~\cite{Florensa2017StochasticNN}. The initial value of the parameter $\sigma$ is set to 0.4.

For the discrete MPE environment, all neural networks of the policies were implemented with fully connected networks that have two layers of 64 hidden units. The size of the learning rate was 0.000018, and the discount factor was 0.99. The batch size and maximum episode length were 8 episodes and 240 steps. For the SparseAnt maze task, the network structure is the same as that of the discrete MPE task. The size of the learning rate is 0.001, and the discount factor is 0.99. The batch size and maximum episode length were 30 episodes and 500 steps.

\subsection{Baseline Methods}
The baseline methods used for performance comparisons varied for different tasks. For discrete control tasks, we compared TCPPO with the following baseline methods: (1) DIPG [10], (2) vanilla PPO~\cite{schulman2017proximal}, (3) Noisy-A2C: the noisy network~\cite{plappert2017parameter} variant of A2C~\cite{mnih2016asynchronous}, (4) Div-A2C: A2C with diversity-driven exploration strategy~\cite{hong2018diversity}, (5) RIDE~\cite{raileanu2019ride}, and (6) NovelD~\cite{zhang2021noveld}. DIPG and Div-A2C promote agent exploration by adding a diversity regularization term to the original RL objective function. PPO is an on-policy RL method derived from TRPO, which is a practical algorithm that addresses continuous state-action spaces. Noisy-A2C is a variant of A2C that adopts noisy networks to increase the randomness of the action outputs. RIDE and NovelD design novel intrinsic reward functions for driving deep exploration in the sparse setting.

For hierarchical continuous control tasks, we compared TCHRL with the following baseline methods: (1) PPO, (2) SNN4HRL~\cite{Florensa2017StochasticNN}, (3) DIPG-HRL, and (4) HAC~\cite{levy2019learning}. SNN4HRL is a state-of-the-art sub-goal-based HRL method. DIPG-HRL is a combination of DIPG~\cite{masood2019diversity} and SNN4HRL, where we use the DIPG objective function to train a high-level policy based on pre-trained skills. Similar to our method, SNN4HRL, and DIPG-HRL share the same set of pre-trained low-level skills as TCHRL. However, pre-trained skills are unadaptable in the training processes of SNN4HRL and DIPG-HRL. Instead, TCHRL sets auxiliary MMD distance rewards for low-level skill training to enable efficient and simultaneous learning of high-level policies and low-level skills without using task-specific knowledge. HAC is a standard end-to-end subgoal-based HRL baseline without delicate techniques of subgoal discovery or quantization.

In multi-agent control tasks, TCMAE is compared with the following baseline methods: (1) SAC~\cite{haarnoja2018soft}, (2) QMIX~\cite{rashid2020monotonic}, (3) EMC~\cite{zheng2021episodic}, (4) MAPPO~\cite{yu2022surprising} and (5) IPPO~\cite{yu2022surprising}. QMIX is a novel value-based method that trains decentralized policies of multiagent in a centralized end-to-end learning fashion. EMC is a curiosity-driven exploration method for deep cooperative multi-agent reinforcement learning (MARL). MAPPO is a PPO-based algorithm with centralized value function inputs for MARL. IPPO represents the independent PPO algorithm where each agent has local inputs for both the policy and value function networks. Recently, MAPPO and IPPO were revisited by~\citep{yu2022surprising} and demonstrated that they could achieve competitive or superior results in various challenging tasks.
\begin{table}[htb]
  \begin{center}
    \caption{Performance comparison with different methods}\label{table:gridworld}
      \begin{tabular*}{0.84\textwidth}{ccccc}
        \toprule
        {} & \multicolumn{2}{@{}c@{}}{$50 \times 50$} &  \multicolumn{2}{@{}c@{}}{$70 \times 70$} \\
        \cmidrule{2-3}\cmidrule{4-5}
        {} &Average reward &Success rate &Average reward &Success rate \\
        \midrule
        TCPPO & \textbf{6.00} & \textbf{1.00} & \textbf{6.00} & \textbf{1.00}\\
        Div-A2C & 4.18 & 0.64 & 3.00 & 0.40\\
        TCPPO w/o ASM & 3.92 & 0.58 & 3.18 & 0.36 \\
        Noisy-A2C & 3.08 & 0.42 & 3.00 & 0.40\\
        DIPG & 3.08 & 0.42 & 1.61 & 0.20\\
        PPO & 2.36 & 0.27 & 2.64 & 0.34\\
        RIDE & \textbf{6.00} & \textbf{1.00} & 5.22 & 0.87\\
        NovelD & 4.50 & 0.75 & 2.25 & 0.38\\
        \bottomrule
      \end{tabular*}
  \end{center}
\end{table}

\section{Evaluation of Results}
\label{sec:experience}
We evaluate the proposed method using several discrete and continuous control tasks. Discrete control tasks comprise 2D grid world environments of different sizes. The continuous control tasks consisted of simulated robotic environments developed by MuJoCo~\cite{todorov2012mujoco}. Furthermore, we directly compared the proposed approach with other state-of-the-art algorithms.

\subsection{Grid-world Task}
\subsubsection{Performance Comparisons with Baseline Methods}
\label{subsec:performance_two_goal_gridworld}
\begin{figure}[!ht]
  \centering
  \subfloat[]{
    \includegraphics[width=7cm]{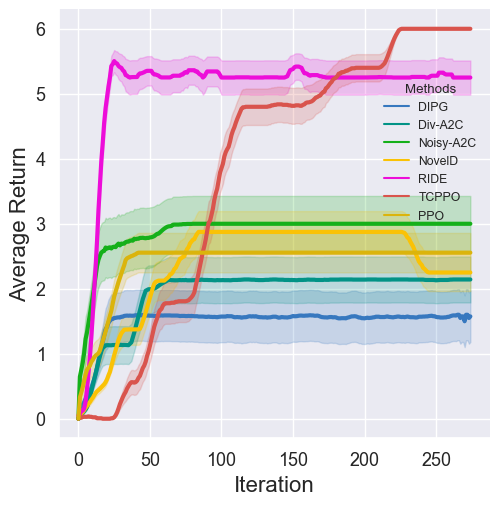}
    \label{fig:maze_return}
  }
  \quad\quad
  \subfloat[]{
    \includegraphics[width=7cm]{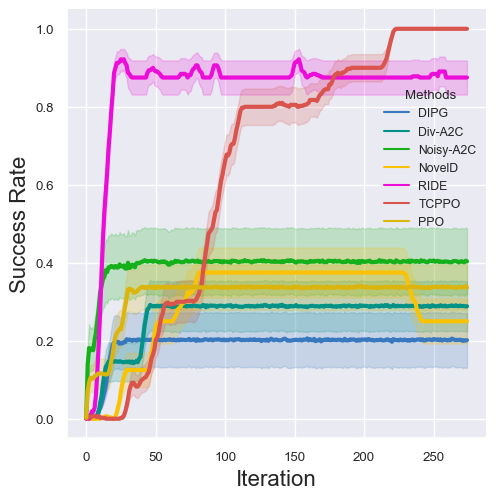}
    \label{fig:maze_rate}
  }
  \caption{
    Learning curves of average return and success rate for different methods in Maze 0 when the replay memory stores previous suboptimal trajectories leading to the local optimal goal. The success rate is used to illustrate the frequency at which the agent reaches the global optimal goal during training processes.
    }
  \label{fig:maze_performance}
\end{figure}

\begin{figure}[ht]
  \centering
  \subfloat[PPO]{
    \label{fig:ppo}
    \includegraphics[width=3.5cm]{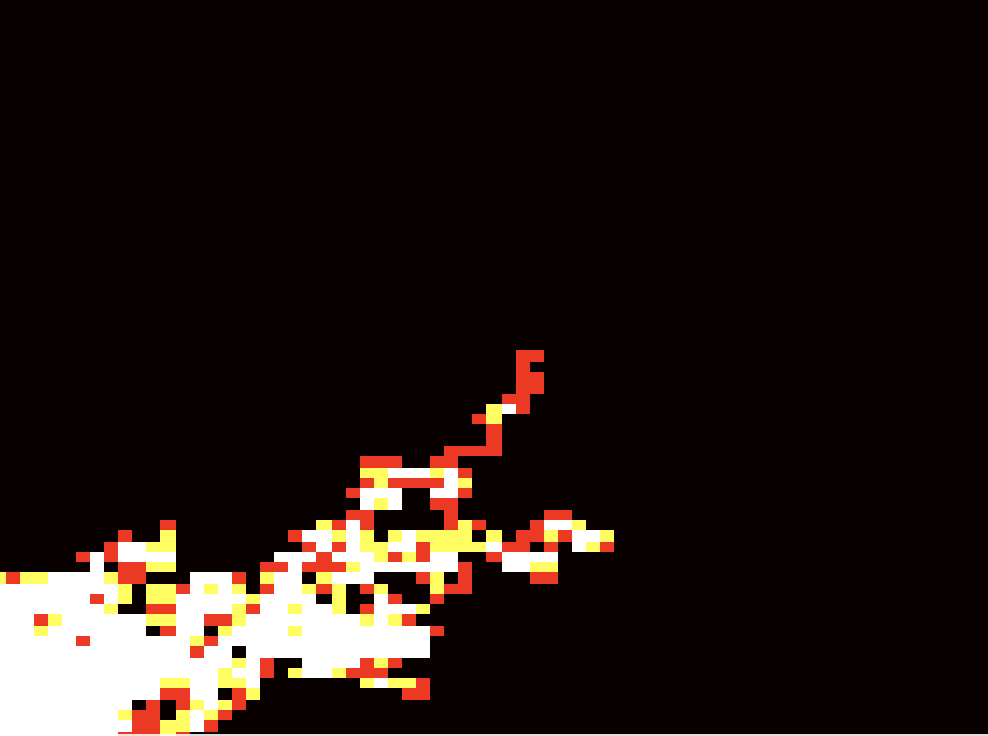}
  }
  \subfloat[Div-A2C]{
    \label{fig:div-ppo}
    \includegraphics[width=3.5cm]{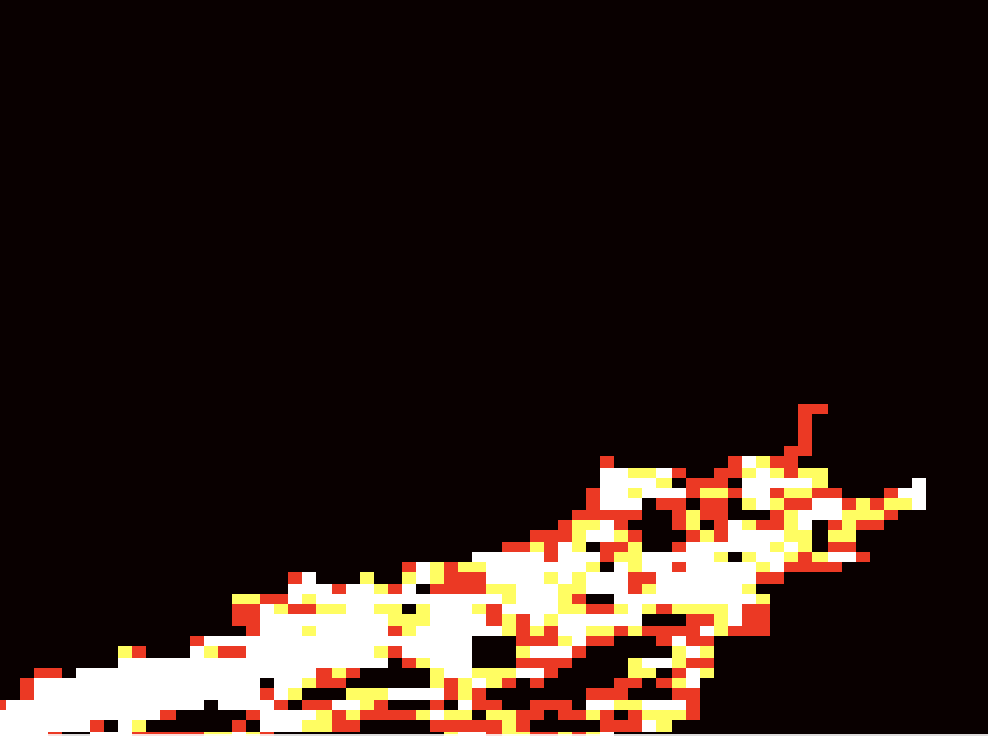}
  }
  \subfloat[DIPG]{
    \label{fig:dipg}
    \includegraphics[width=3.5cm]{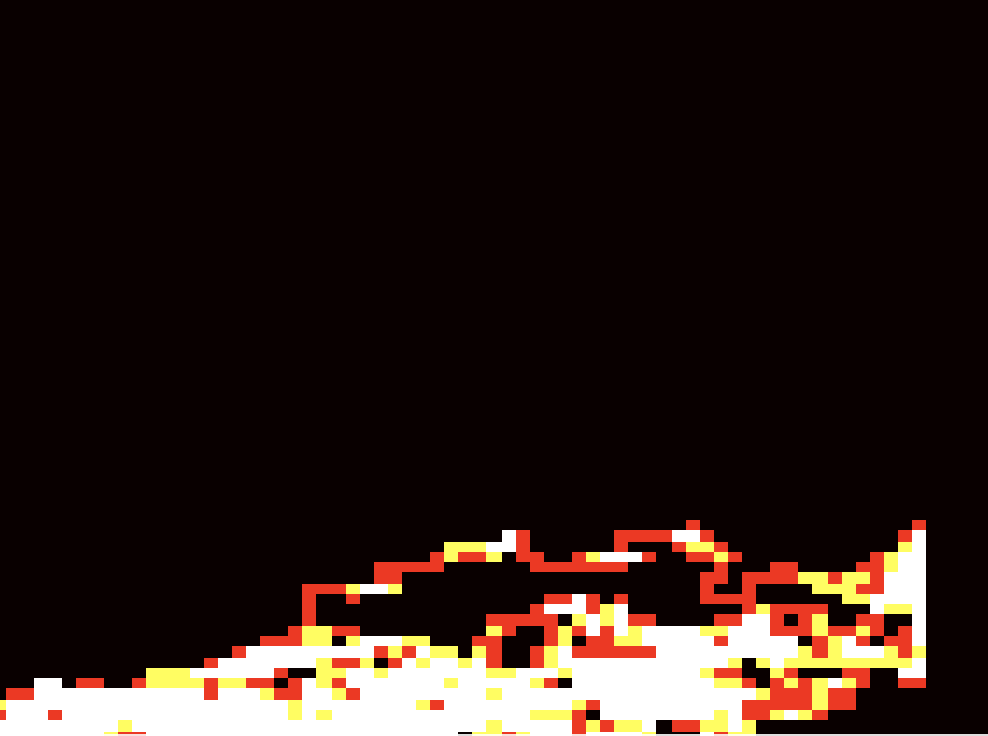}
  }
  \subfloat[RIDE]{
    \label{fig:ride}
    \includegraphics[width=3.5cm]{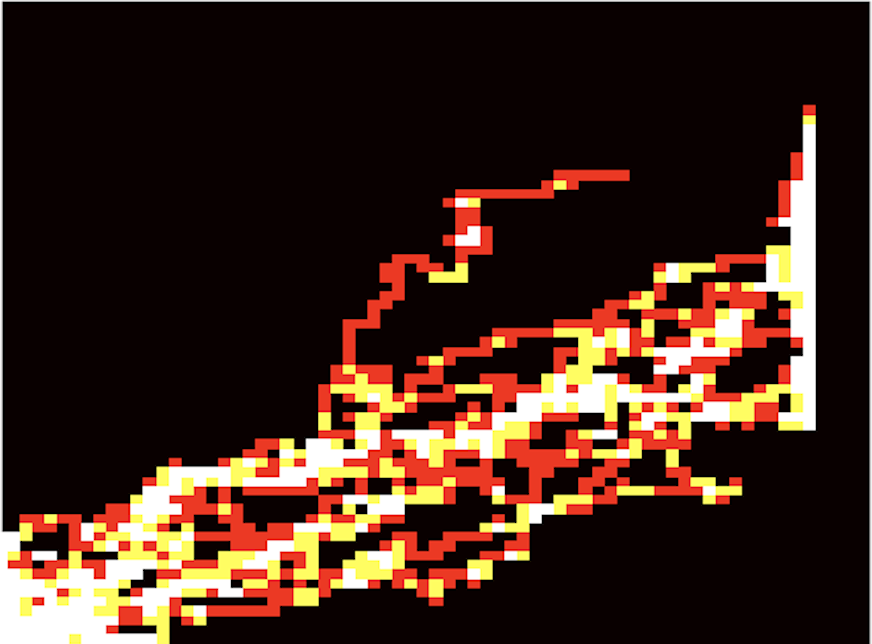}
  }

  \subfloat[Noisy-PPO]{
    \label{fig:noisy}
    \includegraphics[width=3.5cm]{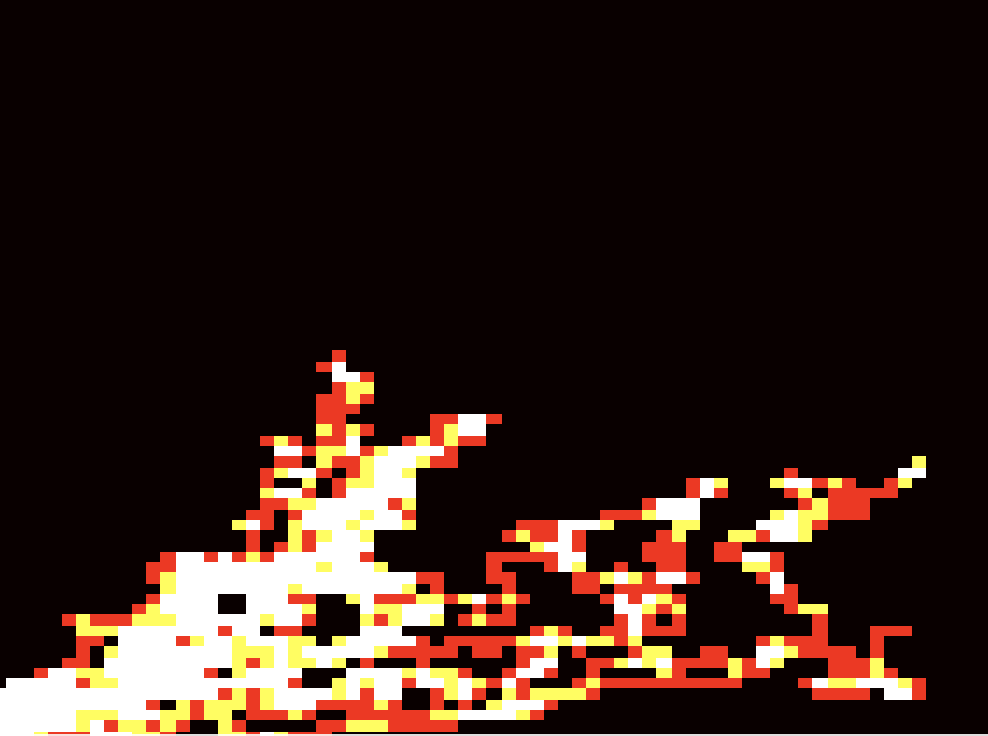}
  }
  \subfloat[NovelD]{
    \label{fig:noveld}
    \includegraphics[width=3.5cm]{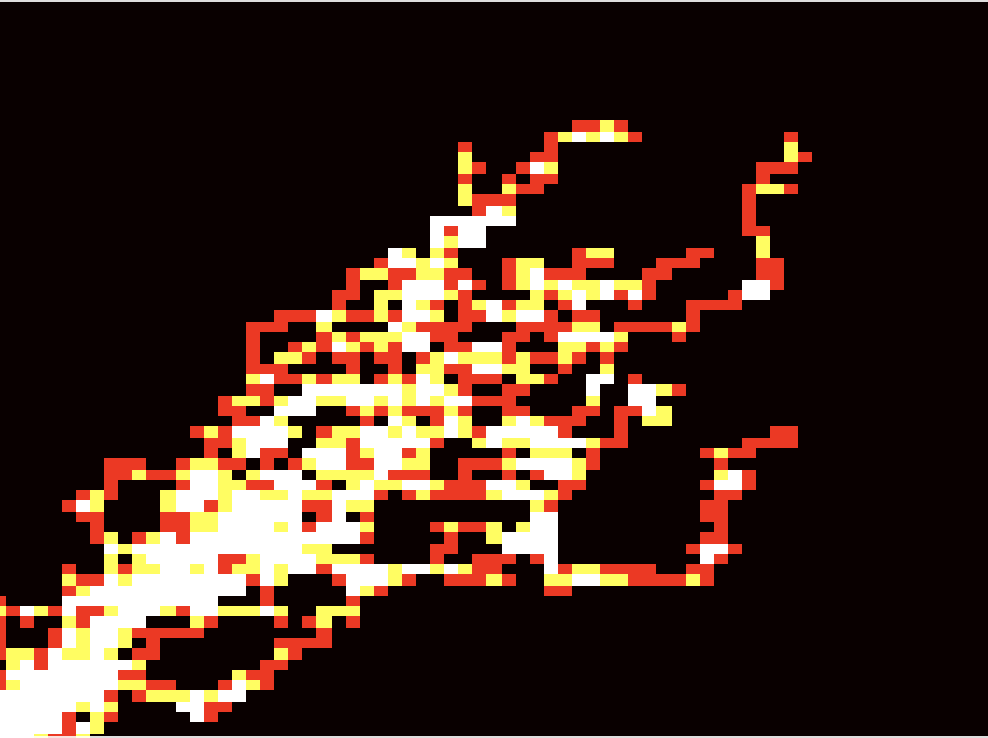}
  }
  \subfloat[TCPPO]{
    \label{fig:tcppo}
    \includegraphics[width=3.5cm]{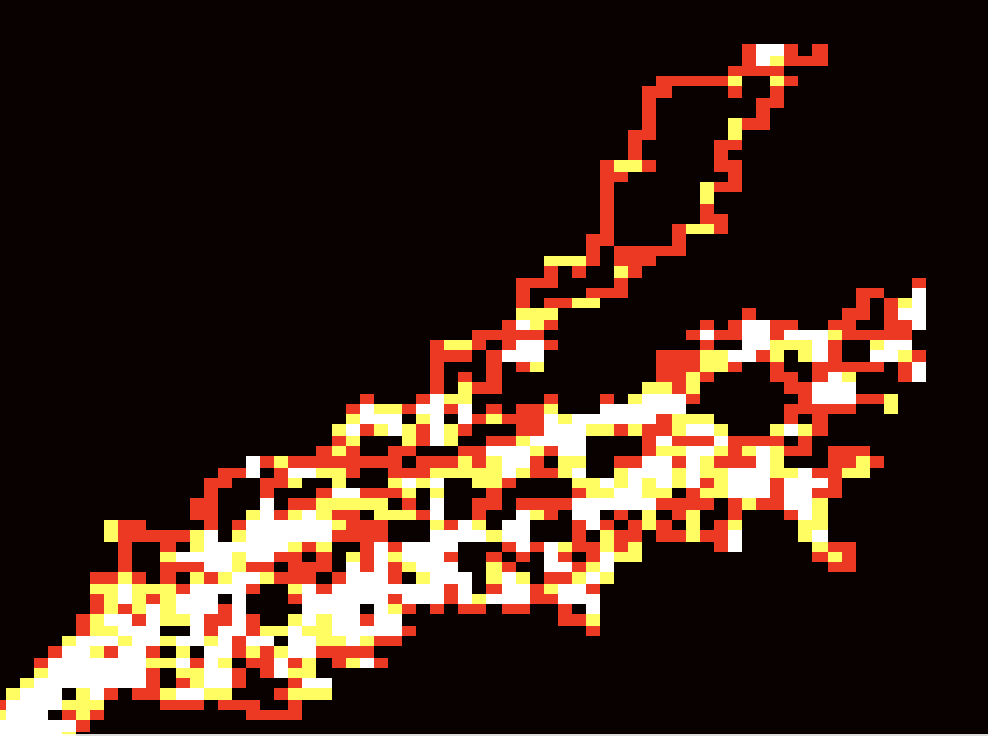}
  }
  \caption{
    State-visitation counts of different algorithms in the $70 \times 70$ grid world: (a) PPO, (b) Div-A2C, (c) DIPG, (d) RIDE, (e) Noisy-A2C, (f) NovelD, and (g) TCPPO.
    }
    \label{fig:state_visitation}
\end{figure}

In this experiment, we combined our trajectory-constrained exploration strategy with the PPO algorithm ~\cite{schulman2017proximal} to obtain the TCPPO algorithm. We evaluated the trajectory-constrained exploration strategy using 2D grid worlds of two different sizes: $50 \times 50$, and $70 \times 70$. The performances of the baseline methods and TCPPO are reported based on their average returns in Table~\ref{table:gridworld}. Some results of the $70\times 70$ maze are presented in Fig.~\ref{fig:maze_performance}. All curves in Fig.~\ref{fig:maze_performance} were obtained by averaging over eleven different random seeds, and for clarity, the shaded error bars represent 0.35 standard errors. Furthermore, we plotted the state-visitation count graphs of each method (Fig.~\ref{fig:state_visitation}) in a $70 \times 70$ grid world, which illustrates the differences in the exploration behaviors for different agents.

As shown in Table~\ref{table:gridworld} and Fig.~\ref{fig:state_visitation}, the results of TCPPO outperform those of the other methods in these two 2D grid worlds of different sizes. Specifically, TCPPO learns faster and achieves higher average returns. The average return of RIDE dramatically increases, and the agent quickly learns to reach the optimal goal. However, the convergent values of both average return and success rate are inferior to those of TCPPO. These results verify that the TCPPO prevents the agent from adopting myopic behaviors and drives deep exploration. From the state-visitation count graph (see Fig.~\ref{fig:state_visitation}), the four baseline approaches visited only a smaller part of the state spaces and hardly collected higher rewards. However, Fig.~\ref{fig:state_visitation}\subref{fig:tcppo} shows that the TCPPO method enhances the exploring behavior of the agent and promotes the agent to explore a wider region of the 2D grid world. Consequently, our method helps the agent escape from the region in which the deceptive reward is located and successfully reach the optimal goal with a higher score. Furthermore, we compare the changing trends in the MMD distances of the different methods in Fig.~\ref{fig:asm+mmd}\subref{fig:mmds}. According to Fig.~\ref{fig:asm+mmd}\subref{fig:mmds}, TCPPO increases the MMD distance between the old and current policies during the training process. Note that DIPG does not learn the optimal policy, but also results in a larger MMD distance, similar to TCPPO.

\subsubsection{Analysis of Adaptive Scaling Method}

The adaptive scaling method for the Lagrange multiplier $\sigma $ serves as a critical component of our proposed trajectory-constrained exploration strategy. This ensures that the agent remains within the feasible region by increasing the Lagrange multiplier if there are state-action pairs outside the feasible region. As shown in Eq.~\eqref{equ:asm}, we also adopt a naive linear decay method to stabilize training processes because it hinders the learning process if the Lagrange multiplier $\sigma$ remains a large constant throughout the whole training process. Table~\ref{table:gridworld} lists the influence of the adaptive scaling method on the agent performance. The experimental results illustrate that the adaptive scaling method enables the agent to gain a higher average return and outperform other methods during training.

\begin{figure}[htb]
  \centering 
  \subfloat[]{
    \label{fig:mmds}
    \includegraphics[width=7cm]{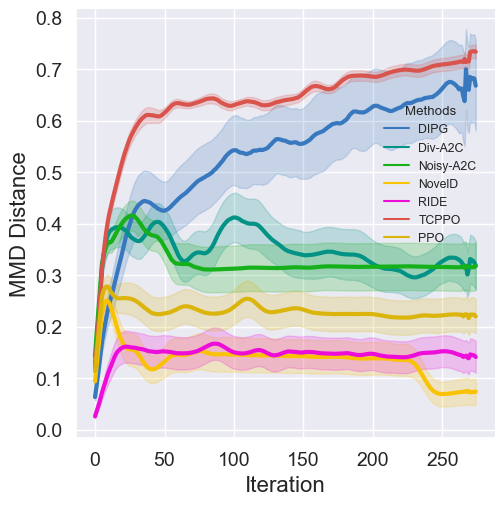}
  }
  \quad\quad
  \subfloat[]{
    \label{fig:sigmatrend}
    \includegraphics[width=7cm]{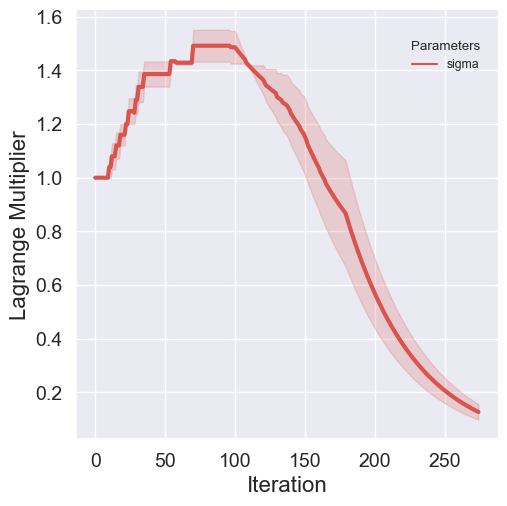}
  }
  \caption{(a) Changing trends in MMD distances; (b) The changing trend of the parameter $\sigma$.}
  \label{fig:asm+mmd}
\end{figure}

Fig.~\ref{fig:asm+mmd}\subref{fig:sigmatrend} shows the changing trend of the Lagrange multiplier $\sigma$ during the training process. At the beginning of the training phase, the agent occasionally encounters a suboptimal goal, which is the same as the goal demonstration trajectories lead to. According to our design in Section~\ref{subsec:asm}, our method drastically increases the value of parameter $\sigma$ to force the agent away from the local maximum. After several episodes, the agent gets rid of the suboptimal goal and gradually explores more diverse regions of the state-action space, wandering between areas of radius $\epsilon$ and $2\epsilon$. After about 200 episodes, the value of the parameter $\sigma$ drops exponentially and slowly, which illustrates that the agent always stays outside the area with a radius of $2\epsilon$. Therefore, our adaptive scaling method significantly improves the efficiency of exploration and protects the agent from falling into a local maximum and adopting myopic behaviors.

\subsection{Deceptive Reacher Task}
\begin{figure}[!ht]
  \centering 
  \subfloat[]{
    \label{fig:reacher_return}
    \includegraphics[width=7cm]{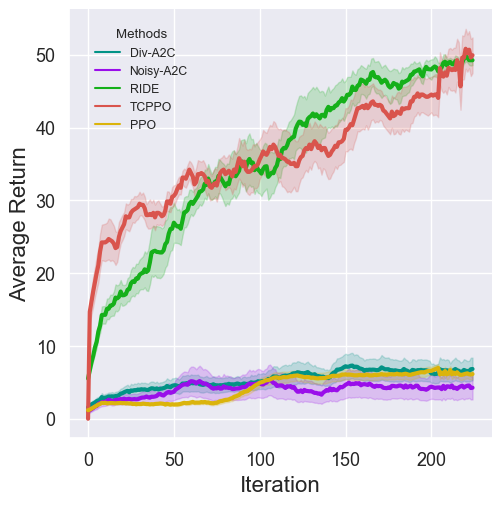}
  }
  \quad\quad
  \subfloat[]{
    \label{fig:reacher_rate}
    \includegraphics[width=7cm]{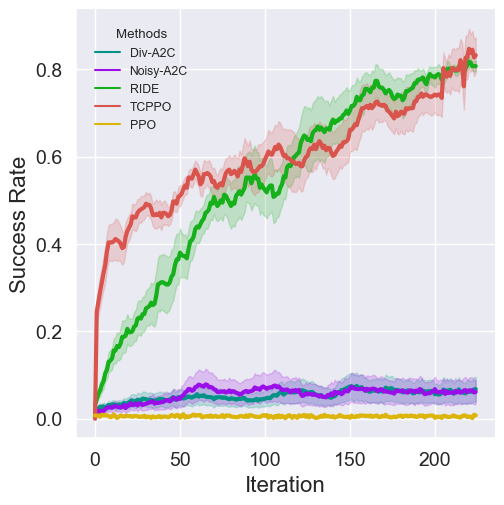}
  }
  \caption{Learning curves of average return and success rate for different methods in the deceptive reacher task.}
  \label{fig:reacher_performance}
\end{figure}

A variant of the classic two-jointed robot arm environment was used to test the proposed method on a continuous robotic control problem. In this deceptive reacher task, we compared our proposed TCPPO algorithm with the same baseline methods used in the grid world task. In this task, to fulfill our design, the replay memory was maintained to store the suboptimal trajectories. Furthermore, we found that it is sufficient to store no more than five trajectories in the replay memory. We report the learning curves of all the methods in Fig.~\ref{fig:reacher_performance} in terms of the average return and success rate. All learning curves were obtained by averaging the results generated with different random seeds, and the shaded error bars represent the standard errors.

Compared with other baseline methods, our approach succeeded in moving out of the box with deceptive rewards and did not adopt myopic behaviors. Therefore, TCPPO can learn faster and achieve higher return values at the end of the training. RIDE fully explored the environment and quickly reached the optimal goal during training. In contrast, the other baseline methods rarely encountered the optimal goal and always fell into the local optimum by collecting deceptive rewards from the box. The PPO is not aimed at long-horizon sparse-reward problems. Consequently, the success rate of the PPO in this continuous control task was always zero. Noisy-A2C and Div-A2C are designed for efficient exploration and occasionally generate trajectories with the optimal rewards. However, our results indicate that these two algorithms do not eliminate the myopic behaviors and cannot learn the optimal policy after training.

\subsection{Performance in Four-Room Maze}
\label{subsec:two-goals}
\subsubsection{Results and Comparisons}
\begin{figure}[!ht]
  \centering
  \subfloat[]{
    \label{fig:exp1_return}
    \includegraphics[width=5cm]{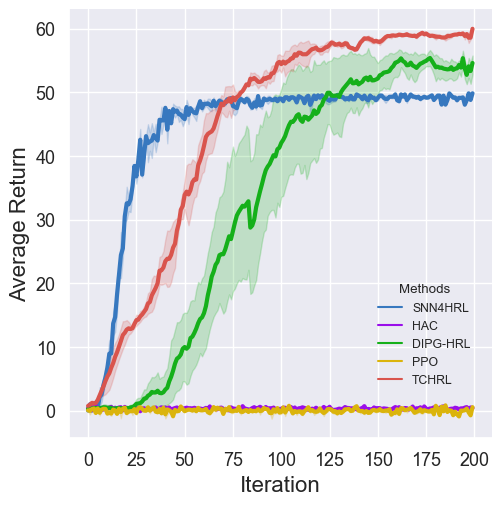}
  }
  \subfloat[]{
    \label{fig:exp1_rate}
    \includegraphics[width=5cm]{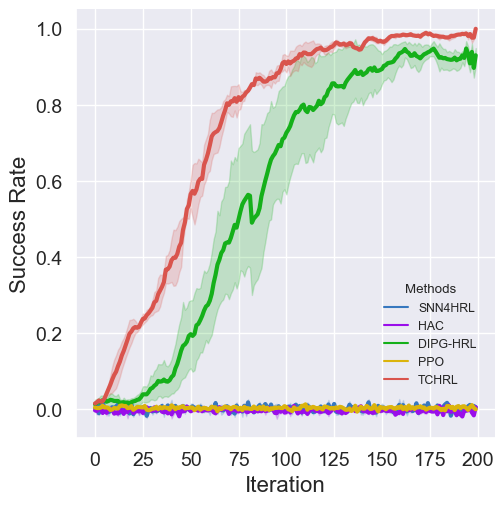}
  }
  \subfloat[]{
    \label{fig:intrinsic_reward}
    \includegraphics[width=5cm]{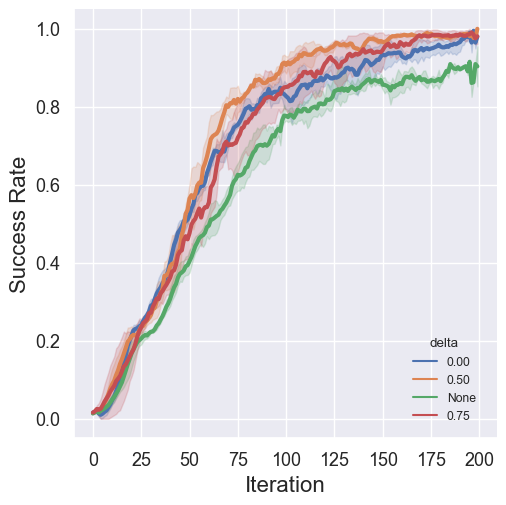}
  }
  \caption{
    (a) Learning curves of average return, (b) Learning curves of success rate. They are obtained when the replay memory stores previous suboptimal trajectories leading to the local optimal goal. The success rate is used to illustrate the frequency at which agents reach the globally optimal goal during the training process. (c) Performance comparison when parameters $\delta$ take different values.}
  \label{fig:exp1}
\end{figure}

In this task, we compared the TCHRL algorithm with the state-of-the-art skill-based method SNN4HRL~\cite{Florensa2017StochasticNN}. In addition, we conducted comparative experiments using a DIPG~\cite{masood2019diversity} variant that is combined with SNN4HRL (denoted DIPG-HRL). In this algorithm, DIPG is only used to train the high-level policy and not to adapt pre-trained skills along with the high-level policy during the training process. We maintain empty replay memories $\mathcal{M}$ for DIPG-HRL and TCHRL at the beginning of the training. When $\mathcal{M}$ is empty, TCHRL and DIPG-HRL degenerate into SNN4HRL. The number of prior suboptimal trajectories is at most $n$, which are generated by the same previous policy. In our experiments, $n=5$ was sufficient to produce satisfactory performance. In the worst case, an agent learns the optimal policy only after it learns all the suboptimal policies and stores all the corresponding trajectories in $\mathcal{M}$. Consequently, the agent may need to sequentially train $g$ different policies in a complete training session, where $g$ represents the number of goals in the maze. For fairness, we trained the PPO agent the same number of times. We plotted the statistical results of different methods based on different goals. All the curves were obtained by averaging over different random seeds, and the shaded error bars represent the confidence intervals. Note that the learning curves shown in Fig.~\ref{fig:exp1} are drawn when the replay memory stores the previous suboptimal trajectories, leading to a suboptimal goal. 

In Fig.~\ref{fig:exp1}, we compare the different methods from two aspects of average return and success rate. Our method is superior to the other methods in both aspects. Specifically, the PPO is not designed for long-horizon tasks with sparse rewards; hence, the success rate of the PPO in achieving the optimal goal is zero in this maze. Although HAC is a subgoal-based HRL algorithm, HAC cannot find goals and obtain any sparse reward according to Fig.~\ref{fig:exp1}, which may be caused by the sensitivity of goal space design~\cite{dwiel2019hierarchical}. SNN4HRL learned only the myopic policy leading to the suboptimal goal and was rewarded with 50 during training, which indicated that the agent was trapped in the local optimum. DIPG-HRL reached the optimal goal at the top-left corner with a high percentage. Moreover, it can be seen from Fig.~\ref{fig:exp1}\subref{fig:exp1_rate} that its success rate in reaching the global optimal goal is lower than that of TCHRL, and it learns slower. Compared with SNN4HRL and DIPG-HRL, the experimental results in Fig~\ref{fig:exp1} demonstrate that TCHRL drives deep exploration and avoids suboptimal and misguided behaviors, thereby attaining a higher score and learning rate. 

\subsubsection{Effect of the Distance Normalization}
%
In this section, we examine the influence of the distance normalization method on the performance of agents in continuous control tasks. We chose different values of the parameter $\delta=0, 0.5, and 0.75$, as the boundary constraints for comparative experiments. The TCHRL algorithm without distance normalization is viewed as a basic comparison baseline, in which the value of the parameter $\delta$ needs to be chosen manually according to different experimental settings. We use ``none'' to represent this method in Fig.~\ref{fig:exp1}\subref{fig:intrinsic_reward}. The experimental results reported in Fig.~\ref{fig:exp1}\subref{fig:intrinsic_reward} demonstrate that by simply setting $\delta=0.5$, TCHRL achieved the best results among the four different parameter settings. Hence, the distance normalization method can reduce the dependence of parameter $\delta$ on the environment and stabilize the learning process.

\subsection{Multiple-Goal Maze Task}
\subsubsection{Performance Comparisons}
\begin{figure}[htb]
  \centering
  \subfloat[]{
    \includegraphics[width=7.cm]{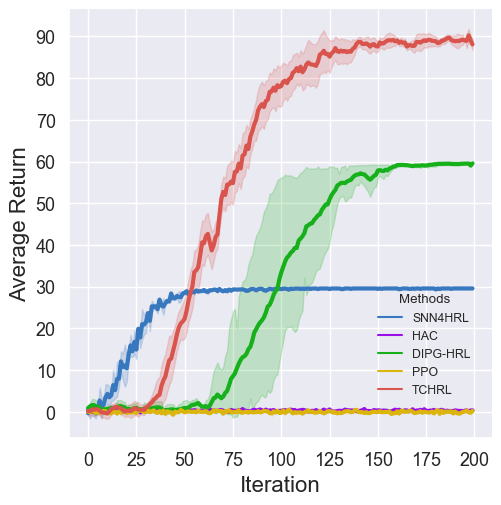}
    \label{fig:exp2_return}
  }
  \quad
  \subfloat[]{
    \includegraphics[width=7.cm]{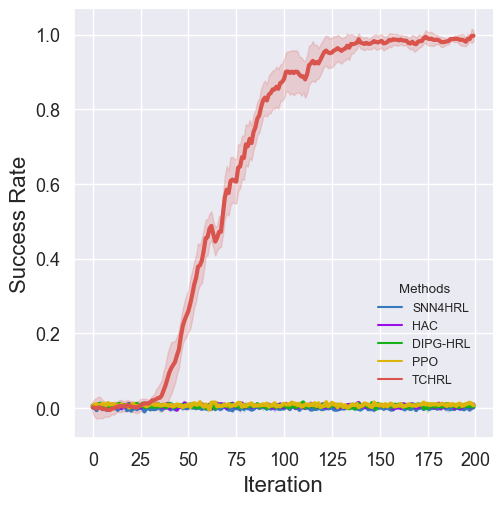}
    \label{fig:exp2_rate}
  }
  \caption{Learning curves of average return and success rate in Maze 1 when prior trajectory 
  buffers store trajectories leading to the local optimal goals.
  The success rate is used to indicate the frequency at which agents reach the globally optimal 
  target during the training process.}
  \label{fig:exp2}
\end{figure}

We compared our algorithm with state-of-the-art skill-based HRL methods SNN4HRL~\cite{Florensa2017StochasticNN} and DIPG-HRL in a multiple-goal maze task. When the replay memory $\mathcal{M}$ maintains trajectories leading to suboptimal goals, TCHRL encourages the agent to generate new trajectories that visit novel regions of the state-action space, gradually expanding the exploration range. As shown in Fig.~\ref{fig:exp2}, the TCHRL agent collects trajectories ending with the optimal goal and learns the policy to collect optimal rewards. In the early training phase, the average return of the TCHRL algorithm was smaller than that of SNN4HRL because our approach preferred to explore the environment more systematically at the beginning of the training time. Moreover, TCHRL gradually adapted its pre-training skills during the training process. Hence, TCHRL does not immediately adopt myopic behaviors to obtain deceptive rewards more easily. All the curves are obtained by averaging over different random seeds, and the shaded error bars represent the confidence intervals.  

In Fig.~\ref{fig:exp2}, the PPO is not an algorithm specifically designed for long-horizon tasks with sparse or deceptive rewards. Therefore, the success rate of the PPO in this maze was zero, as it was in Maze 0. HAC did not receive any reward and learn meaningful policies, which is consistent with previous results in Maze 0. SNN4HRL was rewarded with 30 from the suboptimal blue goal, and DIPG-HRL reached the suboptimal green goal and obtained a reward of 60. Neither received the reward with the highest score from the red goal. Therefore, this result verifies that SNN4HRL and DIPG-HRL cannot fully explore the environment, always adopt myopic behaviors, and learn suboptimal policies. In contrast, our TCHRL can escape suboptimal behaviors and reach the global optimal goal.
The MMD distances between the different trajectories that reached the two suboptimal goals were larger than those of the other two trajectories. Because DIPG uses a regularization term based on trajectory distributions, it is difficult for DIPG to adjust the contributions of the RL objective and regularization term. Therefore, the DIPG-HRL agent tends to learn to reach another suboptimal goal if it has learned a suboptimal policy. Notably, these trajectory distributions were induced by the corresponding policies, and additional trajectory data were required to achieve a good estimation of these distributions. Moreover, DIPG-HRL does not adapt the parameters of the pre-training skills along with a high-level policy, which further degrades the performance of the algorithm. This concept is discussed in detail in the following section. In this manner, we illustrate the inability of the DIPG-HRL agent to learn the optimal policy when the replay memory $\mathcal{M}$ maintains trajectories that lead to two suboptimal goals.

\subsubsection{Adaptation of Pretrained Skills}
\begin{figure}[htb]
  \centering
  \subfloat[]{
    \includegraphics[width=6.cm]{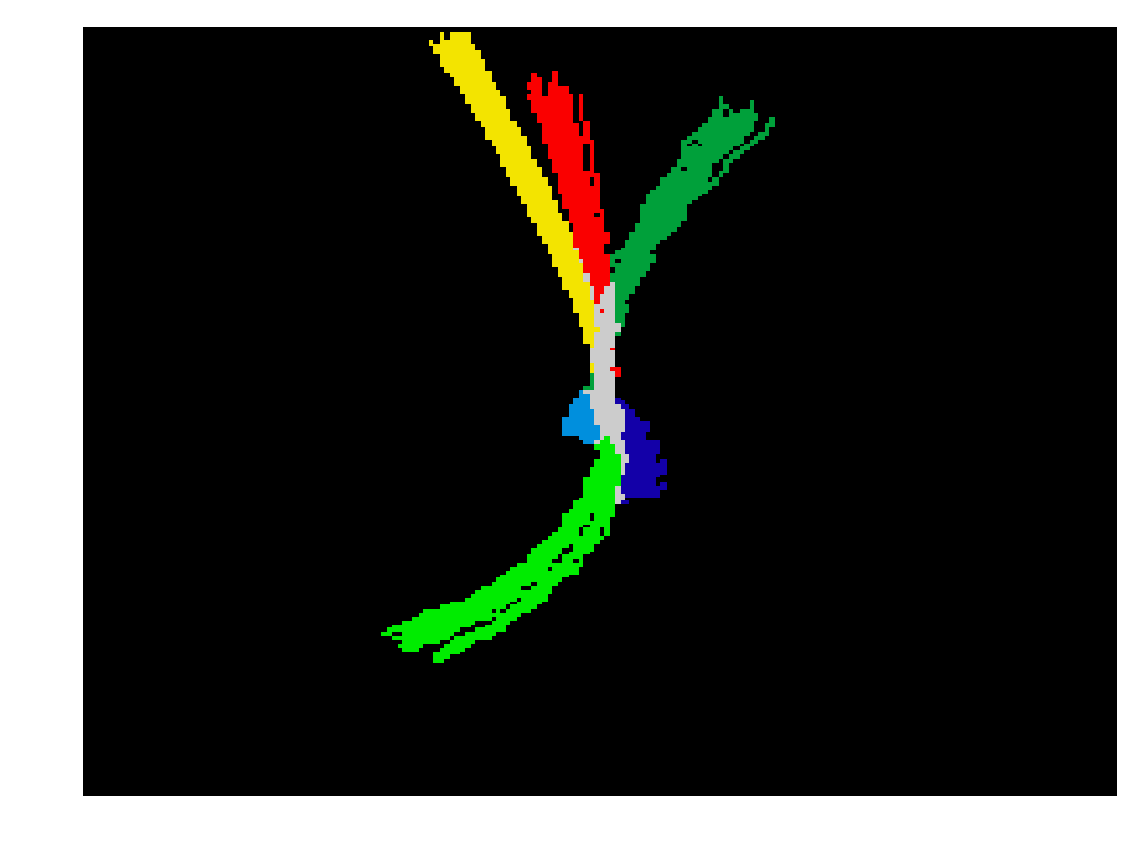}
    \label{fig:skill_before}
  }
  \quad
  \subfloat[]{
    \includegraphics[width=6.cm]{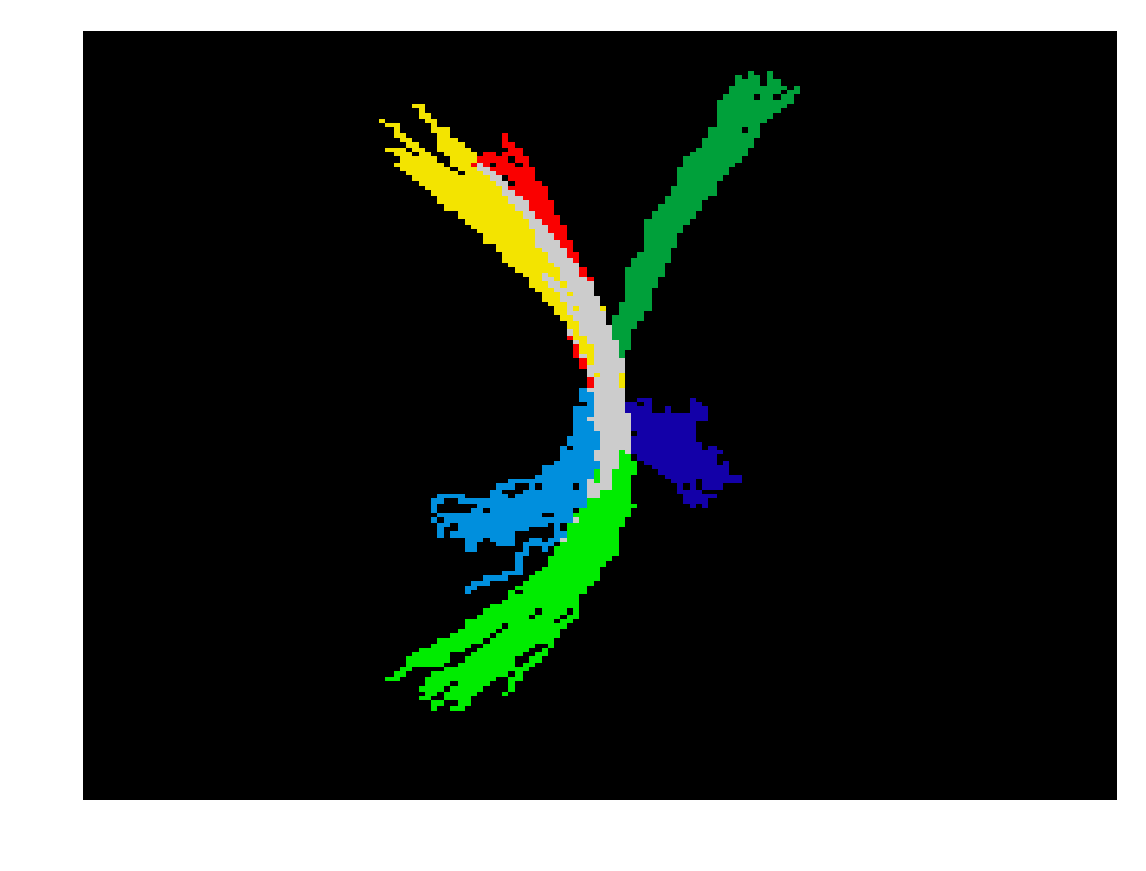}
    \label{fig:skill_after}
  }
  \caption{(a) Visitation graph of the pre-training skills of the agent; (b) Span of skills after training with TCHRL}
  \label{fig:skills_comparison}
\end{figure}

To further explain why TCHRL achieves such excellent performance compared with other baseline methods, we conducted a more in-depth study of the experimental results. TCHRL is a skill-based HRL method. To train the TCHRL agent, we first use stochastic neural networks to learn diverse low-level skills~\cite{Florensa2017StochasticNN}. In Fig.~\ref{fig:skills_comparison}, we compare the low-level skills before and after training in the Swimmer Maze task. The swimmer agent was always initialized at the center of the maze and used a single skill to travel for a fixed number of timesteps, where the colors indicate the latent code sampled at the beginning of the rollout. Each latent code generates a particular interpretable behavior. Given that the initialization orientation of the agent is always the same, different skills are truly distinct ways of moving: forwards, backward, or sideways. 

Many other existing HRL algorithms either use pre-trained skills that are unadaptable or require domain-specific information to define the low-level rewards. In contrast, TCHRL simultaneously adapts low-level skills to downstream tasks while maintaining the generality of low-level reward design by setting auxiliary rewards for low-level skill training, based on the MMD distance. Comparing Fig.~\ref{fig:skills_comparison}\subref{fig:skill_after} with Fig.~\ref{fig:skills_comparison}\subref{fig:skill_before}, we note that the swimmer agent learns to turn right (skill in baby blue) and left (skill in navy blue). It is favorable to perform these two skills in the maze task in Fig.~\ref{fig:mazes}\subref{fig:maze_1} when attempting to reach the green and blue goals. Therefore, the adaptation of pre-trained skills distinctly leads to more diverse skills and effectively drives the agent to explore a wider range of state spaces, which is beneficial for the downstream tasks based on the experimental results.
\begin{figure}[htb]
  \centering
  \subfloat[]{
    \includegraphics[width=7.cm]{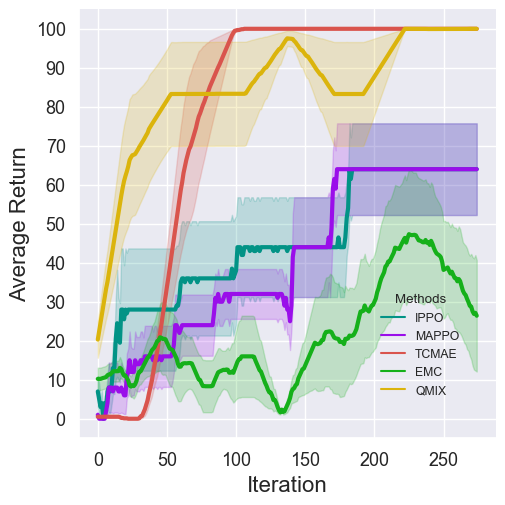}
    \label{fig:ma_grid_return}
  }
  \quad
  \subfloat[]{
    \includegraphics[width=7.cm]{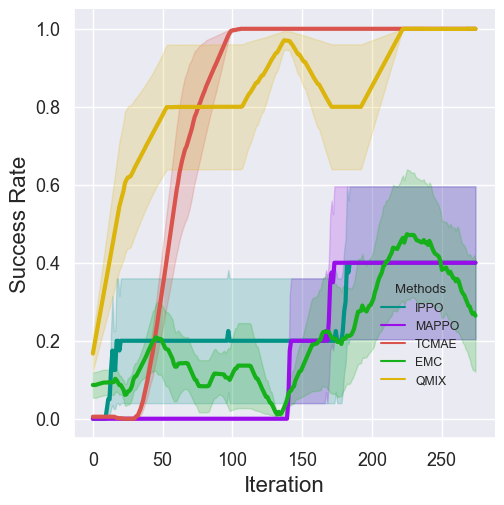}
    \label{fig:ma_grid_rate}
  }
  \caption{Learning curves of average return and success rate for different methods in the discrete MPE environment.}
  \label{fig:ma_grid_performance}
\end{figure}

\subsection{Results of Multi-Agent Tasks}
\subsubsection{Performance Comparisons of Discrete MPE Environment}
In this discrete MPE environment, TCMAE was used to encourage the multi-agent team to fully explore the grid world, and its experimental results were compared with those of the baseline methods. In this task, to fulfill our design, the replay memory of each agent was maintained to store the suboptimal trajectories. The agents of the team shared trajectory information during each epoch. We report the learning curves of all the methods in Fig.~\ref{fig:ma_grid_performance}. All learning curves were obtained by averaging the results generated with different random seeds, and the shaded error bar represents the standard error. Since TCMAE focuses on enhancing the exploration of the multi-agent team, we choose to report the results of agents that generate the highest return values and learning rates.

As shown in Fig.~\ref{fig:ma_grid_performance}, TCMAE achieves competitive and superior performance in terms of average return and success rate compared to the other baseline methods. The large performance gap can be observed from Figs.~\ref{fig:ma_grid_performance}\subref{fig:ma_grid_return} and~\ref{fig:ma_grid_performance}\subref{fig:ma_grid_rate}. Due to exploratory learning behavior, the learning rate of EMC is significantly slower than TCMAE, and the learning process of EMC is more instability than TCMAE. QMIX and IPPO can occasionally find the optimal goal with a simple heuristic exploration strategy, which can lead to inefficient exploration, especially for IPPO. The experimental results demonstrate the great capability of TCMAE to help the multi-agent team explore the environment and reach the optimal goal. 

\subsubsection{Experimental results of SparseAnt Maze Task} 
\begin{figure}[!ht]
  \centering
  \subfloat[]{
    \includegraphics[width=7.cm]{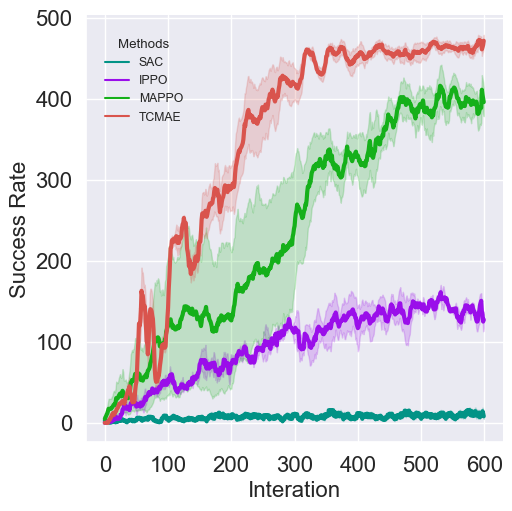}
    \label{fig:ma_ant_return}
  }
  \quad
  \subfloat[]{
    \includegraphics[width=7.cm]{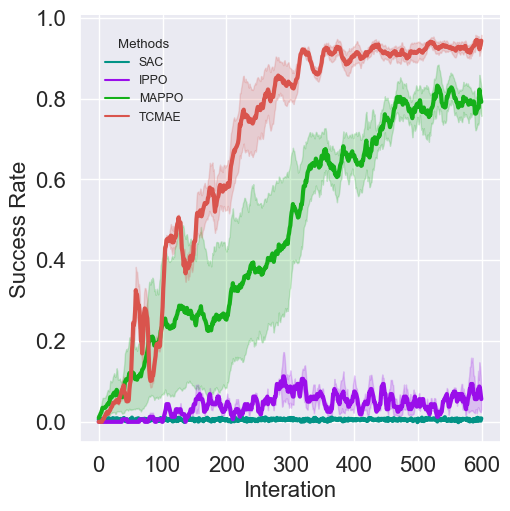}
    \label{fig:ma_ant_rate}
  }
  \caption{Learning curves of average return and success rate for different methods in the SparseAnt maze task.}
  \label{fig:ma_ant_performance}
\end{figure}

To evaluate TCMAE in environments with continuous state-action space, we designed a SparseAnt maze task with sparse and deceptive rewards and compared the performance of TCMAE with several baseline methods. The agent of the team shared the good trajectory information during the training process, which can be used to compute the MMD-based intrinsic reward. We report the learning curves of all the methods in Fig.~\ref{fig:ma_ant_performance}. All learning curves were obtained by averaging the results generated with different random seeds, and the shaded error bar represents the 95\% confidence interval. The available code of QMIX and EMC released by the authors can only be applied to the environments with discrete action spaces; hence they cannot be used as baseline methods in the SparseAnt maze task.

In Fig.~\ref{fig:ma_ant_performance}, the experimental results are reported in terms of average return and success rate. TCMAE achieves a remarkable performance level, and the results of TCMAE considerably outperform baseline methods. ISAC did not obtain significant policy performance improvement after training, and cannot find a policy to collect the optimal rewards. Noticeably, MAPPO can achieve competitive results in both performance metrics, however, its final results are inferior to those of TCMAE and its learning rate is slower than TCMAE. We also compared TCMAE with the multi-agent RL algorithm IPPO to further demonstrate the effectiveness of TCMAE in facilitating multi-agent exploration.

\section{Conclusion}
\label{sec:conclusion}
In this study, we present a trajectory-constrained exploration strategy for long-horizon tasks with large state spaces and sparse or deceptive rewards. We propose to promote the agent's exploration by treating incomplete offline demonstration data as references and demonstrate that this goal can be achieved by introducing an effective distance metric to measure the disparity between different trajectories. We reformulated the policy optimization for RL as a constrained optimization problem, which enhanced the agent's exploration behavior and avoided sensitive hyperparameters. Subsequently, we developed a novel policy-gradient-based algorithm with adaptive clipped trajectory-based distance rewards. Our method can be effectively combined with non-hierarchical and hierarchical RL methods and can continuously adapt pre-trained skills along with high-level policies when the agent employs a hierarchical policy. Furthermore, we introduced an adaptive scaling method and a distance normalization strategy to achieve better performance. The proposed trajectory-constrained strategy is evaluated in large 2D grid worlds and MuJoCo maze environments, and the experimental results show that our method outperformed other baseline algorithms in terms of improving exploration efficiency in large state spaces and avoiding local optima.

Our method encourages agents to visit underexplored regions by considering imperfect offline demonstrations as references. However, when offline imperfect trajectories are on the way to the optimal goal, our method may ignore the fact that exploiting such experiences can indirectly drive deep exploration. Further research may focus on considering the diverse exploration problem in a teamwork setting and exploiting imperfect demonstrations to indirectly accelerate learning and drive deep exploration.

\appendix
\section{Reproducing Kernel Hilbert Spaces}\footnotetext{This section mainly refers to~\cite{bartlett2008rkhs}.}
\subsection{Reproducing Kernel Hilbert Spaces}
\begin{definition}
  An inner product $\langle\mu, \upsilon\rangle$ can be 
  \begin{enumerate}
    \item a dot product: $\langle\mu, \upsilon\rangle = \upsilon^\prime \mu = \sum_{i} \upsilon_i\mu_i;$
    \item a kernel product: $\langle \mu, \upsilon \rangle = k(\upsilon, \mu) = \psi(\upsilon)^\prime \psi(\mu)$ (where $\psi(\mu)$ may have infinite dimensions).
  \end{enumerate}
\end{definition}

Obviously, an inner product $\langle\cdot, \cdot\rangle$ must satisfy the following conditions:
\begin{enumerate}
  \item Symmetry $$\langle\mu, \upsilon\rangle = \langle\upsilon, \mu\rangle\ \forall\mu, \upsilon\in\mathcal{X}$$
  \item Bilinearity $$\langle\alpha\mu + \beta\upsilon, \omega\rangle = \alpha\langle\mu, \omega\rangle + \beta\langle\upsilon, \omega\rangle\ \forall\mu, \upsilon, \omega\in\mathcal{X}, \ \forall\alpha, \beta\in\mathbb{R}$$
  \item Positive definiteness $$\langle\mu, \mu\rangle\ge0,\ \forall\mu\in\mathcal{X}$$
  $$\langle\mu, \mu\rangle=0 \Longleftrightarrow \mu=0$$
\end{enumerate}
\begin{definition}
  A Hilbert Space is an inner product space that is complete and separable with respect to the norm defined by the inner product.
\end{definition}

The vector space $\mathbb{R}^n$ with the vector dot product $\langle a, b \rangle = b^\prime a\ \forall a, b\in\mathbb{R}^n$ is an example of Hilbert space.
\begin{definition}
  $k: \mathcal{X}\times\mathcal{X}\rightarrow\mathbb{R}$ is a kernel if 
  \begin{enumerate}
    \item $k$ is symmetric: $k(x, y) = k(y, x);$
    \item $k$ is positive semi-definite, i.e., $\forall x_1, x_2, \dots, x_n \in\mathcal{X}$, the Gram Matrix $K$ defined as $K_{ij} = k(x_i, x_j)$ is positive semi-definite.
  \end{enumerate}
\end{definition}
\begin{definition}
  $k(\cdot, \cdot)$ is a reproducing kernel of a Hilbert space $\mathcal{H}$ if $\forall f\in\mathcal{H}, f(x) = \langle k(x, \cdot), f(\cdot)\rangle$.
\end{definition}
\begin{definition}
A Reproducing Kernel Hilbert Space (RKHS) is a Hilbert space $\mathcal{H}$ with a reproducing kernel whose span is dense in $\mathcal{H}$. 
\end{definition}
Therefore, an RKHS is a Hilbert space of functions with all evaluation functionals bounded and linear.

\subsection{Build a Reproducing Kernel Hilbert Space}
Given a kernel $k$, we can define a reproducing kernel feature map $\Phi: \mathcal{X}\rightarrow\mathbb{R}^{\mathcal{X}}$ as:
\begin{equation}
  \Phi(x) = k(\cdot, x).
\end{equation}
Consider the vector space:
\begin{equation}
  \label{equ:span}
  \operatorname{span}\left(\left\{\Phi(x: x\in\mathcal{X})\right\}\right) = \left\{\sum_{i=1}^{n}\alpha_i k(\cdot, x_i): n\in\mathbb{N},x_i\in\mathcal{X},\alpha_i\in\mathbb{R}\right\}.
\end{equation}
For $f = \sum_i\alpha_i k(\cdot, \mu_i)$ and $g = \sum_i\beta_i k(\cdot, \upsilon_i)$, define $\langle f,g\rangle=\sum_{i,j}\alpha_i\beta_j k(\mu_i, \upsilon_j)$. Note that: 
\begin{equation}
  \langle f, k(\cdot, x)\rangle  = \sum_i \alpha_i k(x, \mu_i) = f(x),
\end{equation}
i.e., $k$ has the reproducing property.

We show that $\langle f, g\rangle$ is an inner product on the vector space defined by Eq.~\eqref{equ:span} by checking the following conditions:
\begin{enumerate}
  \item Symmetry: $\langle f, g\rangle = \sum_{i,j}\alpha_i\beta_j k(\mu_i, \upsilon_j) = \sum_{i,j}\beta_j\alpha_i k(\upsilon_j, \mu_i) = \langle g, f\rangle;$
  \item Bilinearity: $\langle f, g\rangle = \sum_i \alpha_i g(\upsilon_i) = \sum_j \beta_j f(\mu_j);$
  \item Positive definiteness: $\langle f, f\rangle = \alpha^\prime K\alpha \ge 0$ with equality if $f=0$.
\end{enumerate}
Then we can define a new Hilbert space by completing the inner product space $\langle\cdot, \cdot\rangle$:
\begin{definition}
  For a (compact) $\mathcal{X} \subseteq \mathcal{R}^d$, and a Hilbert space $\mathcal{H}$ of functions $f: \mathcal{X} \rightarrow \mathbb{R}$, we say $\mathcal{H}$ is a \emph{Reproducing Kernel Hilbert Space} if $\exists k: \mathcal{X} \rightarrow \mathbb{R}$, s.t.
  \begin{enumerate}
    \item $k$ has the reproducing property, i.e., $f(x) = \langle f(\cdot), k(\cdot, x)\rangle;$
    \item $k$ spans $\mathcal{H} = \overline{\operatorname{span}\{k(\cdot, x): x\in \mathcal{X}\}}$.
  \end{enumerate}
\end{definition}

\section{Proof of Lemma 1}
\label{sec:app_3}

\mmdgradient*

\begin{proof} 
  Let $x = (s, {a})$ denote a state-action pair. Let $R_i(s, a)$ be the intrinsic reward function derived from the maximum mean discrepancy:
  \begin{equation*}
    R_i(s, a) = \min\left\{D_{\rm MMD}(x, \mathcal{M}) - \delta, 0\right\},
  \end{equation*}
  and $Q_i(\cdot, \cdot)$ be the $Q$-function calculated using $R_i(s, a)$ as the reward:
  \begin{equation*}
    Q_i(s_t, a_t) = \mathbb{E}\left[\sum_{l=0}^{T-t}\gamma^{l}R_i(s_{t+l}, a_{t+l})\right].
  \end{equation*}
  We can easily derive the formula from the policy gradient theorem~\cite{sutton1999policy}:
  \begin{equation*}
    \begin{aligned}
      \nabla_{\theta} D_{\rm MMD} = \mathbb{E}_{\rho_{\pi}(s, a)}\left[\nabla_{\theta}\log\pi_{\theta}(a \vert s)Q_i(s, a)\right].
    \end{aligned}
  \end{equation*}
\end{proof}

\section{Proof of Theorem 1 and Two Relevant Corollaries}
The following lemmas are proved in~\cite{achiam2017constrained}, and we have excerpted them here. The detailed proof process can be found in the appendix of~\cite{achiam2017constrained}.

\begin{lemma} 
  For any function $f: S\to\mathbb{R}$ and any policy $\pi$,
  \begin{equation}
    (1-\gamma) \underset{s \sim \rho_0}{\mathbb{E}}\left[f(s)\right] + \underset{\begin{subarray}{c} s \sim d^{\pi} \\ a \sim \pi \\ s' \sim P\end{subarray}}{\mathbb{E}} \left[ \gamma f(s') \right] - \underset{s \sim d^{\pi}}{\mathbb{E}}\left[f(s)\right] = 0, 
  \end{equation}
  where $\gamma$ is the discount factor, $\rho_0$ is the starting state distribution, and $P$ is the transition probability function. 
\end{lemma}
\begin{lemma}
  \label{policybound0} 
  For any function $f: S \to \mathbb{R}$ and any policies $\pi'$ and $\pi$, define
  \begin{equation}
    T_{\pi,f} (\pi') \doteq \underset{\begin{subarray}{c} s \sim d^{\pi} \\ a \sim \pi \\ s' \sim P\end{subarray}}{\mathbb{E}} \left[ \left(\frac{\pi'(a|s)}{\pi(a|s)} - 1 \right) \left(R(s,a) + \gamma f(s') - f(s) \right)\right],\label{surrogate}
  \end{equation}
  and $\epsilon_f^{\pi'} \doteq \max_s \left| \mathbb{E}_{\begin{subarray}{c}a \sim \pi', s'\sim P\end{subarray}} [R(s,a) + \gamma f(s') - f(s)] \right|$. Consider the standard RL objective function $J(\pi_{\theta}) = \mathbb{E}_{\tau}\left[\sum_{t=0}^{\infty}\gamma^{t} R(s_{t},a_t)\right]$, the following bounds hold:
  \begin{align}
    J(\pi') - J(\pi) \geq \frac{1}{1-\gamma}\left(T_{\pi,f} (\pi') - 2\epsilon_f^{\pi'} D_{TV} (d^{\pi'} || d^{\pi})\right), \label{bound0} \\
    J(\pi') - J(\pi) \leq \frac{1}{1-\gamma}\left(T_{\pi,f} (\pi') + 2\epsilon_f^{\pi'} D_{TV} (d^{\pi'} || d^{\pi})\right), 
    \label{bound0b}
  \end{align}
  where $D_{TV}$ is the total variational divergence. Furthermore, the bounds are tight (when $\pi' = \pi$, the LHS and RHS are identically zero). Here, $\gamma$ is the discount factor and $d^{\pi}$ is the discounted future state distribution.
\end{lemma}

\begin{lemma}
  \label{divergencebound}
  The divergence between discounted future state visitation distributions, $\|d^{\pi'} - d^{\pi}\|_1$, is bounded by an average divergence of the policies $\pi'$ and $\pi$:
  \begin{equation}
    \|d^{\pi'} - d^{\pi}\|_1 \leq \frac{2\gamma}{1-\gamma} \underset{s \sim d^{\pi}}{\mathbb{E}} \left[ D_{TV} (\pi' || \pi)[s]\right], 
  \end{equation}
  where $D_{TV} (\pi'||\pi)[s] = (1/2) \sum_a |\pi'(a|s) - \pi(a|s)|$ is the total variational divergence at $s$.  
  \end{lemma}

\mainthm*

\begin{proof}

  When the MMD gradient of Lemma~\ref{lemma:1} is integrated with the gradient of $J(\theta)$ to update the parameters of the policy, the final gradient $g_\theta$ for the parameter update of Eq.~\eqref{equ: P_I} can be expressed as:
  \begin{equation}
    g_\theta = \mathbb{E}_{\rho_\pi(s,a)}\left[\nabla_{\theta}\log\pi_{\theta}(a \vert s)\left(Q_e(s, a) + \sigma Q_i(s,a)\right)\right].
  \end{equation}

  Due to the similarities of the forms between the $D_{\rm MMD}$ gradient and the RL gradient of $J(\theta)$, MMD-based constraints $\min\left\{D_{\rm MMD}(x, \mathcal{M})-\delta, 0\right\}$ can be viewed as an intrinsic reward $r(s, a) = \min\left\{D_{\rm MMD}(x, \mathcal{M})-\delta, 0\right\}$ for each state-action pair and be integrated with environmental rewards as follows:
  \begin{equation*}
    Q_e(s_t, a_t) + \sigma Q_i(s_t,a_t) = \mathbb{E}_{s_{t+1}, a_{t+1}, \dots}\left[\sum_{l=0}^{\infty}\gamma^l \left(R_e(s_{t+l}, a_{t+l}) + \sigma R_i(s, a)\right)\right]
  \end{equation*}
  
  Then, with the bounds from Lemma~\ref{policybound0} and bound the divergence $D_{TV} (d^{\pi'} || d^{\pi})$ by Lemma~\ref{divergencebound}, we can easily come to the conclusion.
\end{proof}

Theorem~\ref{the:bound} is similar to Theorem 1 of~\cite{achiam2017constrained}. Different the Theorem 1 in~\cite{achiam2017constrained}, our approach transforms constrained optimization problems into unconstrained problems and considers the long-term effects of constraints for returns. Our proposed theorem can be used to analyze the effectiveness of our approach in improving exploration.

\advbound*

\begin{proof}
  Let $f=\tilde{V}$ in Theorem~\ref{the:bound}, and $\tilde{V}$ is the value function computed with environmental and MMD-based rewards. Then, we can derive this corollary.
\end{proof}

\findsigma*

\begin{proof}
  the TV-divergence and KL-divergence are related by $D_{TV} (p||q) \le \sqrt{D_{KL} (p||q) /2}$~\cite{csiszar2011information}. Combining this inequality with Jensen’s inequality, we obtain:
  \begin{equation}
    \label{eq:tvkl}
    \underset{s \sim d^{\pi}}{\mathbb{E}}\left[D_{TV}(\pi'||\pi)[s]\right] \le \underset{s \sim d^{\pi}}{\mathbb{E}}\left[\sqrt{\frac{1}{2} D_{KL}(\pi'||\pi)[s]}\right]
    \le \sqrt{\frac{1}{2} \underset{s \sim d^{\pi}}{\mathbb{E}}\left[D_{KL}(\pi'||\pi)[s]\right]}.
  \end{equation}

  Further, notice that the advantage $A(s,a)$ can be decomposed as the sum of the environmental advantage $A_e (s,a)$ and the MMD-based advantage $A_{i} (s,a)$, which is expressed by:
  \begin{equation}
    \label{eq:decompose_A}
    A_{\pi} (s,a) = A_e(s,a) + A_{i}(s,a),
  \end{equation} 
  Substituting Eq.~\eqref{eq:tvkl} and~\eqref{eq:decompose_A} into the right-hand side of Eq.~\eqref{equ:bound2}, and suppose that $\pi$ and $\pi^\prime$ satisfy $\underset{s \sim d^{\pi}}{\mathbb{E}}\left[D_{KL}(\pi'||\pi)[s]\right] \le \eta$, we can obtain the following equation:
  \begin{equation}
    \label{eq:derive_sigma}
    \begin{aligned}
      &\frac{1}{1-\gamma} \underset{\begin{subarray}{c} s\sim d^{\pi} \\ a \sim \pi' \end{subarray}}{\mathbb{E}} \left[A(s,a) - \frac{2\gamma \epsilon^{\pi'}}{1-\gamma}  D_{TV} (\pi'||\pi)[s] \right] \\ 
      \ge &\frac{1}{1-\gamma}\left[\underset{\begin{subarray}{c} s \sim d^{\pi} \\ a \sim \pi' \end{subarray}}{\mathbb{E}} \left[ A_e (s,a)\right] + \sigma\underset{\begin{subarray}{c} s \sim d^{\pi} \\ a \sim \pi' \end{subarray}}{\mathbb{E}} \left[ A_{i} (s,a)\right] - \frac{\sqrt{2}\gamma \epsilon^{\pi'}}{1-\gamma} \sqrt{\underset{s \sim d^{\pi}}{\mathbb{E}}\left[D_{KL}(\pi'||\pi)[s]\right]}\right]\\ 
      \ge &\frac{1}{1-\gamma}\left[ \underset{\begin{subarray}{c} s \sim d^{\pi} \\ a \sim \pi' \end{subarray}}{\mathbb{E}} \left[ A_e (s,a)\right] + \sigma\underset{\begin{subarray}{c} s \sim d^{\pi} \\ a \sim \pi' \end{subarray}}{\mathbb{E}} \left[ A_{i} (s,a)\right] - \frac{\sqrt{2\eta}\gamma \epsilon^{\pi^\prime}}{1-\gamma}\right].
    \end{aligned}
  \end{equation}
  Let the last term of Eq.~\eqref{eq:derive_sigma} be greater than $\Delta$, i.e.
  \begin{equation}
    \label{eq:sigma_before}
    \frac{1}{1-\gamma}\left[ \underset{\begin{subarray}{c} s\sim d^{\pi} \\ a\sim\pi^\prime \end{subarray}}{\mathbb{E}} \left[A_e (s,a)\right] + \sigma\underset{\begin{subarray}{c} s \sim d^{\pi} \\ a \sim \pi' \end{subarray}}{\mathbb{E}} \left[A_{i} (s,a)\right] - \frac{\sqrt{2\eta}\gamma \epsilon^{\pi'}}{1-\gamma}\right] \ge \Delta.
  \end{equation}
  After sorting out Eq.~\eqref{eq:sigma_before}, we get:
  \begin{equation}
    \label{eq:sigma_appendix}
    \sigma \ge {\underset{\begin{subarray}{c} s\sim d^{\pi} \\ a\sim\pi' \end{subarray}}{\mathbb{E}}^{-1}\left[A_{i} (s,a)\right]}\left[(1-\gamma)\Delta - \underset{\begin{subarray}{c} s \sim d^{\pi} \\ a \sim\pi^\prime \end{subarray}}{\mathbb{E}}\left[A_e (s,a)\right] + \frac{\sqrt{2\eta} \gamma \epsilon^{\pi^\prime}}{1 -\gamma}\right].
  \end{equation}
  That is to say, $L(\theta^\prime, \sigma) - L(\theta, \sigma) \ge \Delta$ when $\sigma$ satisfies Eq.~\eqref{eq:sigma_appendix}.
\end{proof}

\section{An Overview of Hyperparameter Configurations \& Search Spaces}
\subsection{Stable Baselines Default Configurations}
Table~\ref{table:default_configs} shows the default hyperparameters we used in the experiments of Section~\ref{sec:experience}. TCHRL is based on TRPO in the hierarchical navigation tasks, and hence some hyperparameters are not necessary for it, and we use "-" to indicate this situation.
\begin{table}[ht]
  \centering
  \label{table:default_configs}
  \caption{Hyper-parameter choices of LOPE for different tasks.}
  \begin{center}
  \begin{small}
  \begin{tabular}{lcccr}
  \toprule
  \bf{Hyper-parameter} & \bf{Grid World} &\bf{Hierarchical} & \bf{Discrete MPE} & \bf{SparseAnt} \\
  \midrule
  Value Function Learning Rate & $0.00012$ & $-$ & $0.00012$ & $0.001$ \\
  Policy Learning Rate & $0.000018$ & $0.01$ & $0.000018$ & $0.00009$ \\
  Action Regularization & None & True & None  & True \\
  Policy Optimizer & Adam & Adam & Adam & Adam \\
  Value Function Optimizer & Adam & Adam & Adam & Adam \\
  Max Length per Episode & $240$ & $1e4$ & $240$ & $500$ \\ 
  Episodes per iteration & $8$ & $16$ & $8$ & $30$ \\
  Epoches per iteration & $65$ & $-$ & $65$ & $30$ \\
  Discount Factor & $0.99$ & $0.99$ & $0.99$ & $0.99$ \\
  Gradient Clipping & False & False & False & False \\
  Clipped Epsilon & $0.20$ & -  & $0.20$ & $0.24$ \\ 
  Initial Sigma & $0.42$ & $0.60$ & $0.42$ & $0.70$ \\ 
  Delta & $0.7$ & $0.7$ & $0.7$ & $0.7$ \\
  \bottomrule
  \end{tabular}
  \end{small}
  \end{center}
\end{table}

\subsection{Sweep Values of Hyperparameters}
To determine the optimal hyperparameter value, we first swept over the different hyperparameter values for the same dimension in a relatively large range. For some hyperparameters, we further refined the scope of the search.

For the learning rate, the first search scope was $\{1e-2,5e-3,1e-3,5e-4,1e-4,5e-5,1e-5,5e-6,1e-6,5e-7\}$. Then, we narrowed down the search and reduced the search step size, and the second search scope was $\{5e-5,4e-5,3e-5,2e-5,1e-5,9e-6,8e-6,7e-6,6e-6,5e-6\}$. After two rounds of searching, we find that the highest performance of TACE was obtained when $\texttt{learn rate}=2e-5$. Finally, we fine-tuned its value around $2e-5$ and finally obtained the learning rate of $1.8e-5$.

For the clip range hyperparameter of PPO, which was the basis of our algorithm TCPPO, we selected its proper value by scanning the values in $\{0.0,0.1,0.2,0.3,0.4,0.5,0.6,0.7,0.8,0.9\}$. We selected the initial Lagrange multiplier $\sigma$ from the search space $\{0.0,0.1,0.2,0.3,0.4,0.5,0.6,0.7,0.8,0.9\}$. The suitable value of $\epsilon$ is determined by scanning over the set $\{0.0, 0.05, 0.1, 0.15, 0.2, 0.25, 0.3, 0.35, 0.4\}$. 

Because we adopted the heuristic adaptive scaling method, the value of $\delta$ is environment-independent, and hence the optimal value of $\delta$ can be selected for different environments by sweeping over the set $\{0.0,0.1,0.2,0.3,0.4,0.5,0.6,0.7,0.8,0.9\}$.

\section{Pseudocode of TCPPO Algorithms}
Algorithm~\ref{algo:tcppo} describes our method TCPPO in detail. 

\textbf{Notations:}

$\theta =\,$ Policy parameters

$\sigma =\,$ Lagrange multiplier

$\alpha =\,$ Learning rate

$N =\,$ Size of on-policy buffer

$G =\,$ Number of goals in the environment

\begin{algorithm}[H]
  \caption{Trajectory-Constrained PPO}\label{algo:tcppo}
  \begin{algorithmic}[1]

    \STATE $\mathcal{B}\leftarrow\,$ on-policy buffer
    \STATE $\mathcal{M}\leftarrow\,$ empty replay memory
    \STATE  $\theta, \sigma\sim\,$ initial parameters
    
    \FOR{$i = 0$ to $G$}
    \FOR{$j = 0$ to $N$}
    \STATE Generate batch of $N$ trajectories $\{\tau_i\}_{i}^{n}$ and store them in $\mathcal{B}$

    \STATE Calculate the MMD distance $D_{\rm MMD}(x, \mathcal{M})$ for each state-action pair $x$

    \STATE Normalize the MMD distance $D_{\rm MMD}(x, \mathcal{M})$ according to Eq.~\eqref{equ:normal}

    \STATE Estimate the MMD gradient $\nabla_{\theta}D_{\rm MMD}$ using $\mathcal{M}$ and $\mathcal{B}$

    \STATE Estimate the policy gradient $\nabla_{\theta}J$ based on $\mathcal{B}$

    \STATE Calculate the final gradient $\nabla_{\theta}L = \nabla_{\theta} J + \sigma \nabla_\theta D_{\rm MMD}$

    \STATE $\theta \leftarrow \theta + \alpha\nabla_{\theta}L$

    \STATE Update $\sigma$ according to Eq.~\eqref{equ:asm}

    \ENDFOR
    \ENDFOR
  \end{algorithmic}
\end{algorithm}

\section{Training Process of TCHRL Algorithms}
Algorithm~\ref{algo:tchrl} describes our method TCHRL in detail. At each time step, the algorithm is executed according to the framework shown in Fig.~\ref{fig:execution}. State-action pairs generated by the current policy are stored in the on-policy set $\mathcal{B}$. Then we use these experiences to estimate the policy gradient $\nabla_{\theta}L$ according to Eq.~\eqref{equ: P_I} and calculate the diversity measurement $D_{\rm MMD}$ between different policies. Finally, we update parameters of $\pi_{\theta}$ with the gradient ascent algorithm and adapt the penalty factor according to Eq.~\eqref{equ:asm}.

\textbf{Notations:}

$\theta := \{\theta_h, \theta_l\} =\,$ Policy parameters

$\sigma =\,$ Lagrange multiplier

$\alpha =\,$ Learning rate

$N =\,$ Size of on-policy buffer

$G =\,$ Number of goals in the environment

\begin{algorithm}[H]
  \caption{Trajectory-Constrained HRL}
  \label{algo:tchrl}
  \begin{algorithmic}[1]
    \STATE Pre-train low-level skills $\pi_{\theta_l}$
    \STATE Initialize the on-policy buffer $\mathcal{B}$
    \STATE Initialize the replay memory $\mathcal{M}$
    \STATE Initialize high-level policy network parameters $\theta_h$
    \STATE Initialize the parameter $\sigma$
    \FOR{$i = 0$ to $M$}
    \FOR{$j = 0$ to $N$}
    \STATE Generate batch of $N$ trajectories $\{\tau_i\}_{i}^{n}$ and store them in $\mathcal{B}$

    \STATE Calculate the normalized MMD distance $D_{\rm MMD}^h$ and $D_{\rm MMD}^l$ for the high-level policy and the low-level policy respectively

    \STATE Normalize the MMD distance $D_{\rm MMD}^h$ and $D_{\rm MMD}^l$ separately according to Eq.~\eqref{equ:normal}

    \STATE Estimate the MMD gradient $\nabla_{\theta}D_{\rm MMD}$ according to~\eqref{equ:nabla_mmd_hierarchical_appx} using the estimations of $D_{\rm MMD}^h$ and $D_{\rm MMD}^l$

    \STATE Estimate the policy gradient $\nabla_{\theta}J$ based on $\mathcal{B}$

    \STATE Calculate the final gradient $\nabla_{\theta}L = \nabla_{\theta}J + \sigma\nabla_\theta D_{\rm MMD}$

    \STATE $\theta \leftarrow \theta + \alpha\nabla_{\theta}L$

    \STATE Update $\sigma$ according to Eq.~\eqref{equ:asm}

    \ENDFOR
    \ENDFOR
  \end{algorithmic}
\end{algorithm}

\section{TCHRL Learning Frameworks}
\label{subsec:TCHRL}
In this section, we introduce the implementation of our exploration strategy using a hierarchical policy. Our implementation was based on the state-of-the-art skill-based hierarchical reinforcement learning algorithm SNN4HRL~\cite{Florensa2017StochasticNN}. Instead of freezing the low-level skills during the training phase of the downstream task, our proposed diversity incentive adapts the pre-training skills along with the high-level policy training.

\begin{figure}[htb]
  \centering
  \includegraphics[scale=0.41]{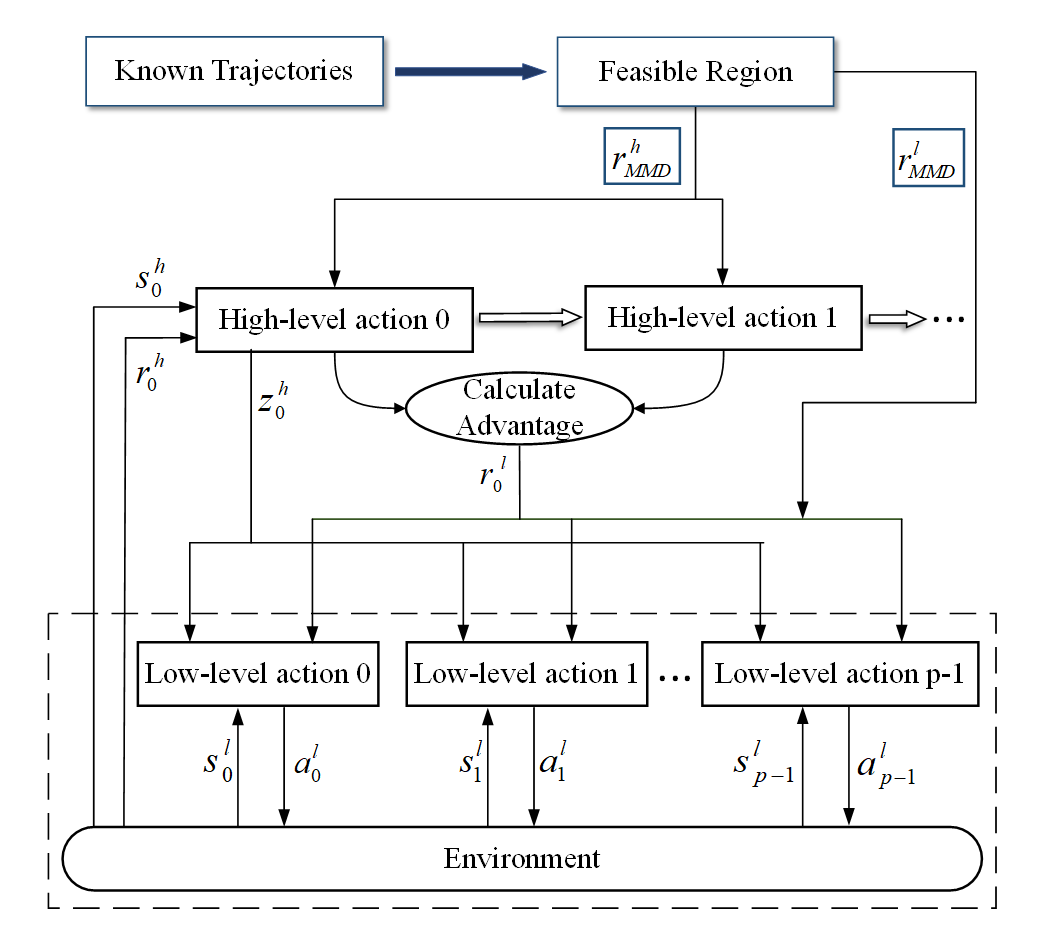}
  \caption{
    Graphical description of the algorithm execution process. When the TCHRL algorithm first starts, the high-level policy outputs a latent code $z_0$, then the low-level policy inputs state along with this latent code to the neural network. After the low-level policy takes $p$ time steps, the high-level policy selects another latent code $z_1$.
    }
  \label{fig:execution}
\end{figure}

Fig.~\ref{fig:execution} shows the execution process of our trajectory-constrained hierarchical reinforcement learning algorithm. In this HRL algorithm with a 2-level hierarchy, the agent takes a high-level action (or a latent code) $ z_t $ every $p$ timesteps after receiving a new observation $s_t$, i.e., $ z_t = z_{kp}$  if $ kp \le t \le (k+1)p-1$. The low-level policy $ \pi_{\theta_l}$ is another neural network that treats the current observation $s_t$ and high-level action $z_{t}$ as inputs, and its outputs are low-level actions $a_{t}$ used to interact with the environment directly. The skills selected by the high-level policy are executed by the low-level policy for the next $p$ time steps. In our framework. Different skills of the low-level policy are distinguished by different latent codes $z$, and a single stochastic neural network~\cite{tang2013learning} is employed to encode all the pre-trained skills~\cite{Florensa2017StochasticNN,li2019hierarchical}.

Under our framework, a trajectory $ \tau $ can be expressed as:
\begin{equation}
  \label{equ:tau}
  \tau = (s_0, a_0, s_1, a_1, \dots, s_{T}, a_{T}),
\end{equation}
and the probability of generating this trajectory can be expressed as~\cite{Li2020SubpolicyAF}:
\begin{equation}
  \begin{aligned}
    \label{equ: p_tau}
    p(\tau) = \left(\prod_{k=0}^{T/p} \left[\sum_{j=1}^{m}\pi_{\theta_h}(z_j \vert s_{kp}) \prod_{t=kp}^{(k+1)p - 1}\pi_{\theta_l}(a_t \vert s_t, z_j) \right] \right) 
    \cdot \left[\rho(s_0)\prod_{t=1}^{T}P(s_{t+1} \vert s_t, a_t)\right],
  \end{aligned}
\end{equation}
where $m$ represents the number of different low-level skills, $s_0\sim\rho_0(s_0)$, $z_{kp}\sim\pi_{\theta_h}(z_{kp} \vert s_{kp})$, $a_t \sim \pi_{\theta_l}(a_t \vert s_t, z_t)$, and $s_{t+1} \sim P(s_{t+1} \vert s_t, a_t)$. Hence, we define the sequence of a high-level action $z_k$ followed by $p$ low-level actions $(a_{kp}, \dots, a_{(k+1)p - 1})$ as a macro action $\tilde{a}$, then the probability of a macro action 
can be written as
\begin{equation}
  \label{equ:pi_macro}
  \pi(\tilde{a} \vert s_{kp}) = \pi_{\theta_h}(z_j \vert s_{kp})\prod_{t=kp}^{(k+1)p-1}\pi_{\theta_l}(a_t \vert s_t, z_j).
\end{equation}

TCHRL allows a high-level policy to select a pre-trained skill to perform low-level actions over several time steps. Unlike general sub-goal-based HRL methods~\cite{levy2019learning,nachum2018data,nachum2018near}, these pre-trained skills do not need to reach the sub-goals set by a high-level policy. Moreover, these low-level skills can be rewarded by environmental and MMD distance rewards, which can be used to compute the Q-functions $\tilde{Q}_h(s_{kp}, z_{kp})$ and $\tilde{Q}_l(s_{t}, z_{kp}, a_t)$ for low-level and high-level policies, respectively. By Lemma~\ref{lemma:1}, combining Eqs.~\eqref{equ: p_tau} and~\eqref{equ:pi_macro}, the following hierarchical MMD gradient formula holds:
\begin{equation}
  \label{equ:nabla_mmd_hierarchical_appx}
  \begin{aligned}
    \nabla_{\theta} D_{\rm MMD}=\mathbb{E}_{\rho_\pi(s, \tilde{a})}
    \Bigg[\nabla_{\theta_h}\log\pi_{\theta_h}(z_{kp} \vert s_{kp})\tilde{Q}_h(s_{kp}, z_{kp})
    +\sum_{t=kp}^{(k+1)p - 1}\nabla_{\theta_l}\log\pi_{\theta_l}(a_t \vert s_t, z_{kp})\tilde{Q}_l
    (s_{t}, z_{kp}, a_t)\Bigg],
  \end{aligned}
\end{equation}
where $\tilde{Q}(s_{kp}, z_{kp})$ is calculated with high-level trajectories based on Eq.~\eqref{equ:Q}, and $\tilde{Q}(s_{t}, z_{kp}, a_t)$ is calculated with low-level trajectories similar to $\tilde{Q}(s_{kp}, z_{kp})$.

In this framework, replay memory $\mathcal{M}$ is maintained to store the suboptimal trajectories generated by past suboptimal policies learned during the past training process. Instead, state-action pairs collected by the current policy are stored in the on-policy buffer $\mathcal{B}$. Our method uses these trajectories in $\mathcal{M}$ and $\mathcal{B}$ to compute the MMD distance.

\section{Training Process of TCMAE Algorithms}
Algorithm~\ref{algo:tcmae} describes our method TCMAE in detail. 

\textbf{Notations:}

$I := \{1, 2,\dots, n\}=$ the finite sets of agents

$\theta := \{\theta_h, \theta_l\} =\,$ Policy parameters

$\sigma =\,$ Lagrange multiplier

$\alpha =\,$ Learning rate

$N =\,$ Size of on-policy buffer

$G =\,$ Number of goals in the environment

\begin{algorithm}[H]
  \caption{Trajectory-Constrained MAE}
  \label{algo:tcmae}
  \begin{algorithmic}[1]
    \STATE Initialize the on-policy buffer $\mathcal{B}$
    \STATE Initialize the replay memory $\mathcal{M}_k, k\in I$
    \STATE Initialize the parameters $\theta_k$ and $\sigma_k, k\in I$

    \FOR{$i = 0$ to $M$}
    \FOR{$j = 0$ to $N$}

    \STATE Generate batch of $N$ trajectories $\{\tau_i\}_{i}^{n}$ and store them in $\mathcal{B}_k$ for each agent

    \FOR{$k\in I$}
    \STATE Update $\mathcal{M}_k$ using the trajectories of local observations and actions generated by the $k-$th agent

    \STATE Calculate the normalized MMD distance $D_{\rm MMD}^k$ for the $k-$th agent using the episodic state-action visitation distributions of $\mathcal{M}_l, l\in I, l\neq k$

    \STATE Normalize the MMD distance $D_{\rm MMD}^k$ separately according to Eq.~\eqref{equ:normal}

    \STATE Estimate the MMD gradient $\nabla_{\theta}D_{\rm MMD}^k$ according to Eq.~\eqref{equ:nabla_E_D_mmd} with the estimations of $D_{\rm MMD}^k$ 

    \ENDFOR

    \STATE Estimate the policy gradient $\nabla_{\theta}J$ based on $\mathcal{B}$

    \STATE Calculate the final gradient $\nabla_{\theta}L = \nabla_{\theta}J + \sum_{k\in I}\sigma_k\nabla_\theta D_{\rm MMD}^k$

    \STATE $\theta \leftarrow \theta + \alpha\nabla_{\theta}L$

    \STATE Update each $\sigma_k$ according to Eq.~\eqref{equ:asm}

    \ENDFOR
    \ENDFOR
  \end{algorithmic}
\end{algorithm}

\section{Additional Experimental Results}
\begin{figure}[htb]
  \centering
  \includegraphics[scale=0.35]{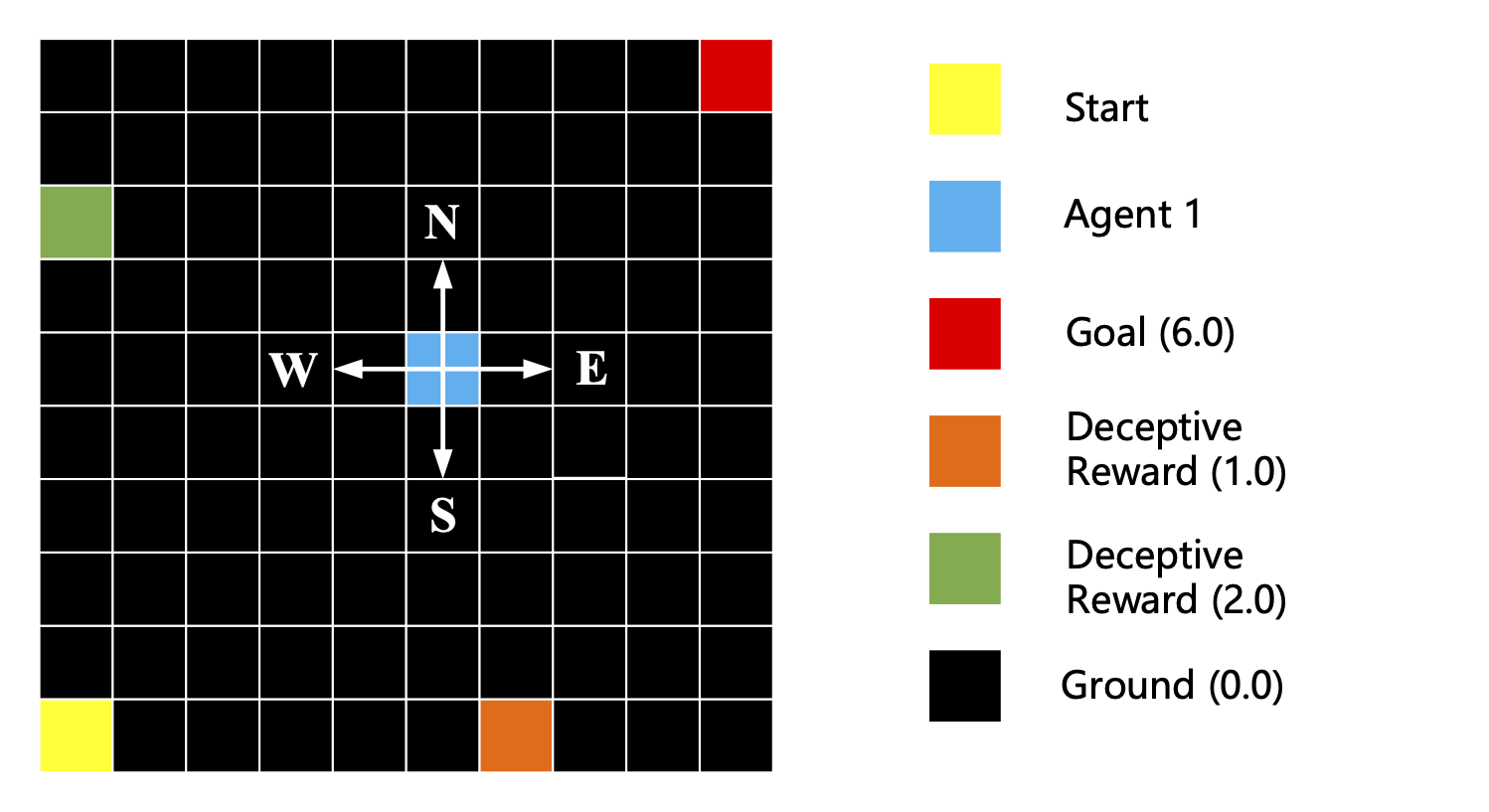}
  \caption{Gridworld with multiple goals.}
  \label{fig:three_goal_grid}
\end{figure}

In this section, we further evaluate the performance of TCPPO in tasks with discrete state-action spaces, as shown in Fig.~\ref{fig:grid+reacher}\subref{fig:discrete_maze}. Compared with the maze in Fig.~\ref{fig:grid+reacher}\subref{fig:discrete_maze}, the only difference in Fig.~\ref{fig:three_goal_grid} is the addition of a new deception goal with a reward of 2. In addition, the settings for both grid world mazes were identical. The size of the 2D grid world maze was $70\times70$ steps. All curves are obtained by averaging over different random seeds, and for clarity, the shaded error bars represent 0.35 standard errors.

\begin{figure}[!ht]
  \centering
  \subfloat[]{
    \includegraphics[width=7cm]{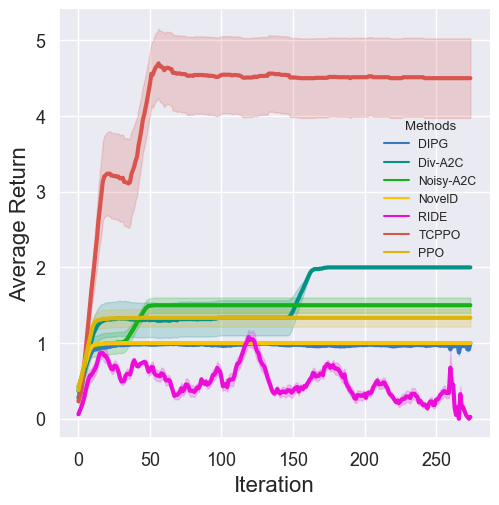}
    \label{fig:three_goal_return}
  }
  \quad\quad
  \subfloat[]{
    \includegraphics[width=7cm]{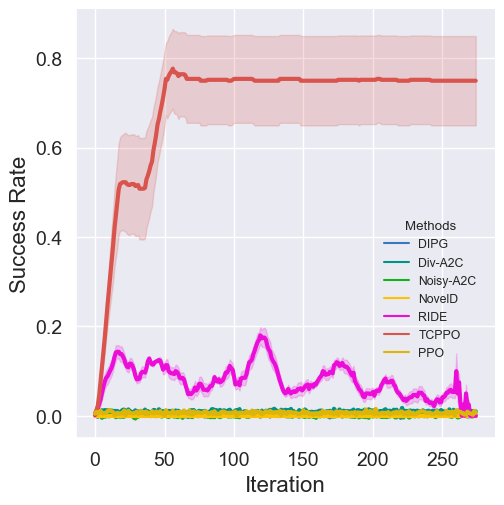}
    \label{fig:three_goal_rate}
  }
  \caption{
    Learning curves of average return and success rate for different methods when the replay memory stores previous suboptimal trajectories leading to the two local optimal goals. The success rate is used to illustrate the frequency at which the agent reaches the global optimal goal during training processes.
    }
  \label{fig:three_goal_performance}
\end{figure}

As shown in Figs.~\ref{fig:three_goal_performance}, the TCPPO algorithm significantly outperformed the other baseline methods in this 2D grid world maze. Similar to the results in Section~\ref{subsec:performance_two_goal_gridworld}, TCPPO learned faster and achieved higher average returns. Meanwhile, with the help of TCPPO, the agent avoided adopting myopic behaviors and performed deep exploration. In particular, the performance gap between our approach and the baseline methods was even greater than that described in Section~\ref{subsec:performance_two_goal_gridworld}. Hence, the proposed method has obvious advantages for this more difficult task.

\begin{figure}[!ht]
  \centering
  \includegraphics[width=8cm]{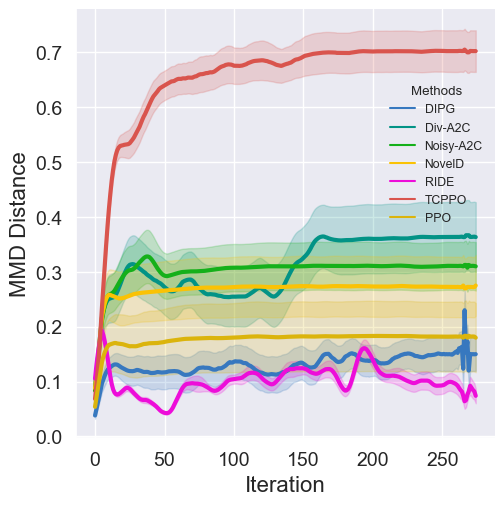}
  \caption{Changing trends of MMD distances.}
  \label{fig:three_goal_mmd}
\end{figure}

Furthermore, we compared the changing trends in the MMD distances of the different methods, as shown in Fig.~\ref{fig:three_goal_mmd}. The results illustrate that TCPPO increases the MMD distance between the old and current policies during the training process. By contrast, the four baseline methods could not achieve such a large MMD distance.

\section*{Acknowledgment}
This work was supported by the National Key R\&D Program of China (2022ZD0116401)
\section*{Conflict of interest} 
The authors declare that they have no known competing financial interests or personal relationships that could have appeared to influence the work reported in this paper
\section*{Authors’ Contributions}
\textbf{Guojian Wang:} Conceptualization, Methodology, Software, Formal analysis, Writing–Original Draft. \textbf{Faguo Wu:} Writing–Review and Editing, Supervision. \textbf{Xiao Zhang:} Validation, Supervision, Funding acquisition. \textbf{Ning Guo:} Formal analysis, Visualization. \textbf{Zhiming Zheng:} Supervision, Validation, Funding acquisition.
\section*{Availability of data and materials}
The datasets generated during and/or analysed during the current study are available from the 
corresponding author on reasonable request. Code for this paper is available at \url{https://github.com/buaawgj/TACE}.

\bibliographystyle{elsarticle-num}
\bibliography{reference}

\end{document}